# INTERPRETABLE MACHINE LEARNING PREDICTS PARKINSON'S DISEASE SEVERITY USING MOTION-CORRECTED QSM MRI AND MULTIBAND MULTI-ECHO FMRI FEATURES

*Aixa X. Andrade*[1,2,3]

AFFILIATIONS
[1]Lyda Hill Department of Bioinformatics
[2]Department of Biomedical Engineering
[3]University of Texas Southwestern Medical Center, Dallas, Texas, USA

## Abstract

**Introduction:** Objective neuroimaging biomarkers may improve Parkinson's disease motor assessment by capturing brain variation not directly observable from clinical examination. We used interpretable machine learning to predict current motor severity, measured by MDS-UPDRS Part III, from susceptibility-sensitive QSM and multiband multi-echo resting-state fMRI-derived ReHo features.

**Methods:** Regional QSM and ReHo features were extracted from 28 participants, including 24 individuals with Parkinson's disease and 4 controls. Thirteen feature-set experiments evaluated imaging-only, clinical-only, imaging-plus-clinical, full, reduced, and multimodal inputs. Support Vector Regression, Elastic Net, Random Forest, and XGBoost models were trained using nested cross-validation with inner-loop hyperparameter optimization. Performance was assessed using pooled held-out $R^2$, RMSE, MAE, Pearson correlation, permutation testing, and the proportion of participants predicted within ±5 MDS-UPDRS Part III points.

**Results:** Imaging-only models carried meaningful predictive signal, whereas the clinical-only model performed weakly, indicating that predictions were not driven by clinical variables alone. Full fMRI, full QSM, and clinical variables provided the strongest global fit, explaining 45.4% of variance in motor severity. In contrast, selected QSM plus clinical variables produced the most clinically close predictions, with 75.0% of participants predicted within ±5 points and the lowest MAE among top-performing models. SHAP highlighted biologically plausible regions, including cerebellar, thalamic, striatal, insular, and motor cortical features.

**Conclusion:** QSM and multiband multi-echo fMRI-derived ReHo capture distinct, interpretable dimensions of Parkinson's disease motor severity. Rather than identifying a single best biomarker, these findings show that structural and functional imaging contribute differently depending on the clinical prediction goal in this cohort.

# Introduction

Parkinson's Disease (PD) is a neurodegenerative disorder characterized by sequelae including both motor and non-motor symptoms[1]. Here, we focus on motor severity because it is central to clinical evaluation and commonly quantified with the Movement Disorder Society Unified Parkinson's Disease Rating Scale (MDS-UPDRS), with Part III assessing clinician-rated motor examination severity[2]. PD involves dopaminergic neuronal loss in the substantia nigra pars compacta[3], Lewy body pathology, and dysregulated iron metabolism[4]. Clinical scales capture observable symptoms but provide limited direct insight into the neurobiological variation underlying motor impairment.

Structural and functional MRI can complement clinical assessment by capturing disease-relevant brain variation. Quantitative Susceptibility Mapping (QSM) estimates tissue magnetic susceptibility and is particularly sensitive to iron-related susceptibility changes[5]. This makes QSM relevant for Parkinson's disease, where altered iron deposition has been implicated in motor-related regions, including the substantia nigra, red nucleus, and motor cortex, and where QSM enables non-invasive in vivo quantification of iron-sensitive susceptibility variation[6]. Recent QSM studies have revealed elevated susceptibility in PD versus healthy controls and showed significant correlations between regional susceptibility and multiple clinical measures[7], supporting QSM as a structural marker of PD-relevant tissue changes.

Functional MRI complements QSM by capturing dynamic blood-oxygen-level-dependent (BOLD) activity rather than susceptibility-related tissue variation. Multiband multi-echo resting-state fMRI enables faster sampling of BOLD signal[8], multi-echo combination, biological signal isolation, and denoising[9, 10]. In this study, we derived Regional Homogeneity (ReHo) features from multiband multi-echo resting-state fMRI. These properties are relevant for Regional Homogeneity (ReHo), a commonly used approach for reducing 4D resting-state fMRI data into lower-dimensional measures of local functional synchronization[11]. Because ReHo depends on the similarity of local BOLD signal fluctuations, improved temporal sampling and signal quality may support more robust regional ReHo estimates[12]. Together, QSM and multiband multi-echo fMRI-derived ReHo provide complementary structural and functional perspectives on PD-relevant brain variation.

Machine-learning studies in Parkinson's disease have shown that diverse clinical measures, either alone or combined with neuroimaging features, can predict clinical severity [13-16], motor fluctuations[17] or diagnostic status[18]. However, prior work has not directly evaluated the combined value of QSM and resting-state fMRI for predicting current motor examination severity. Prior severity-prediction studies using resting-state fMRI-derived ReHo have focused mainly on current and future MDS-UPDRS total scores when combined with clinical variables[16, 19], while other fMRI-based studies have focused on UPDRS-III or MDS-UPDRS Part III motor scores[20-22]. In contrast, multimodal or connectivity-based approaches have often emphasized progression, prognosis, or subtype differentiation rather than current motor severity prediction from complementary structural and functional MRI biomarkers[23, 24].

The key innovation of the present study is the integration and systematic comparison of susceptibility-sensitive QSM and multiband multi-echo fMRI-derived ReHo features for predicting current clinician-rated PD motor severity, measured by MDS-UPDRS Part III. Building on prior work, we refine imaging-based severity prediction by: (1) integrating QSM and fMRI-derived ReHo to test whether structural and functional MRI provide non-redundant predictive information; (2) focusing on current MDS-UPDRS Part III rather than total UPDRS scores; (3) comparing imaging-only, clinical-only, imaging-plus-clinical, full, reduced, and multimodal feature sets; and (4) using SHAP[25] to identify biologically plausible features driving prediction.

Accordingly, this study was designed to answer several related questions: (1) whether neuroimaging-derived features can predict current clinician-rated motor severity in PD, measured by MDS-UPDRS Part III; (2) whether QSM and multiband multi-echo resting-state fMRI-derived ReHo provide complementary and non-redundant predictive information; (3) whether multimodal models combining QSM and fMRI outperform single-modality models; (4) whether clinical variables improve imaging-based prediction or provide distinct contextual information; (5) whether biologically guided feature reduction improves model performance in this small-sample, high-dimensional setting; (6) whether different model classes vary in their suitability for this prediction task; and (7) which imaging and clinical features most strongly contribute to motor severity prediction and whether these features are neurobiologically plausible.

# Methods

## Participants

All imaging data was acquired at the UT Southwestern Medical Center with full participant consent as part of an IRB- approved Deep Phenotyping Imaging Study. Study inclusion criteria included adults ≥40 years, Participants included either adults with a diagnosis of clinically probable Parkinson's disease (PD) or healthy controls without evidence of neurodegenerative disease. The exclusion criteria included contraindications to MRI or severe comorbidities precluding imaging, major non-PD neurological or psychiatric disorders.

The cohort for this work included data from a subset of the study participants that have successful QSM MRI and MBME fMRI acquisition and MDS-UPDRS part III rating by a Physician Assistant trained and certified on this rating system. This cohort of 28 participants has 24 with PD, including those with newly diagnosed, de novo PD, mild PD, moderate PD, and advanced/severe PD, as well as 4 controls (Table 1). Of the 30 subjects originally available, two were excluded because their MDS-UPDRS Part III scores fell outside the mean ± 3 standard deviation range of the sample (2.6–42.32). For participants with PD, only scans acquired during the ON medication state were included in the analysis.

## Clinical features encoding

General clinical variables were included: sex (male/female), handedness (left/right), age, levodopa equivalent daily dose (LEDD), time since last levodopa dose, and diagnosis. Together, these variables resulted in a total of 6 clinical features. Time since last levodopa dose was included because medication state can influence the observed motor examination[26, 27]. For control participants, *time since dose* was set to a large value corresponding to age in minutes, calculated as age × 365.25 × 24 × 60, to reflect the absence of levodopa treatment. For Parkinson's disease participants assessed in the ON medication state, missing *time since dose* values were imputed using the mean *time since dose* among available PD-ON participants.

**Table 1. Cohort demographics.**

Data from 24 subjects with Parkinson's disease and 4 control subjects were included. Binarized clinical features are reported as counts, and all other numerical variables are reported as mean ± sample standard deviation.

| Variable | Count (N=28) | |
|---|---|---|
| Female sex | 12 (42.9%) | |
| Right handedness | 23 (82.1%) | |
| | **Mean** | |
| Age (years) | 70.57 ± 8.68 | |
| Height (cm) | 173.04 ± 11.10 | |
| Weight (kg) | 77.17 ± 13.54 | |
| | **Mean Control** | **Mean PD** |
| UPDRS part 3 | 13.75 ± 4.99 | 22.46 ± 6.62 |
| LEDD | - | 731.17 ± 266.35 |
| Time since dose (min) | - | 106.83 ± 91.37 |

## QSM and fMRI feature extraction

QSM and resting-state multiband multi-echo fMRI were used to derive regional structural and functional imaging features, respectively, with acquisition, reconstruction, preprocessing, and registration procedures described in Supplementary Methods 1.1–1.2. As illustrated in Figure 1, the QSM images, mean magnetic susceptibility was extracted from 206 bilateral atlas regions defined in a standard QSM atlas[28]. From the preprocessed fMRI data, regional homogeneity (ReHo) was computed to quantify local synchronization of resting-state BOLD signal fluctuations[11] , and mean regional ReHo values were extracted using the Schaefer 2018 atlas with subcortical labels[16, 29]. These procedures yielded 206 QSM features and 135 fMRI-derived ReHo features per participant, representing complementary susceptibility-sensitive structural and local functional measures for prediction of MDS-UPDRS Part III scores.

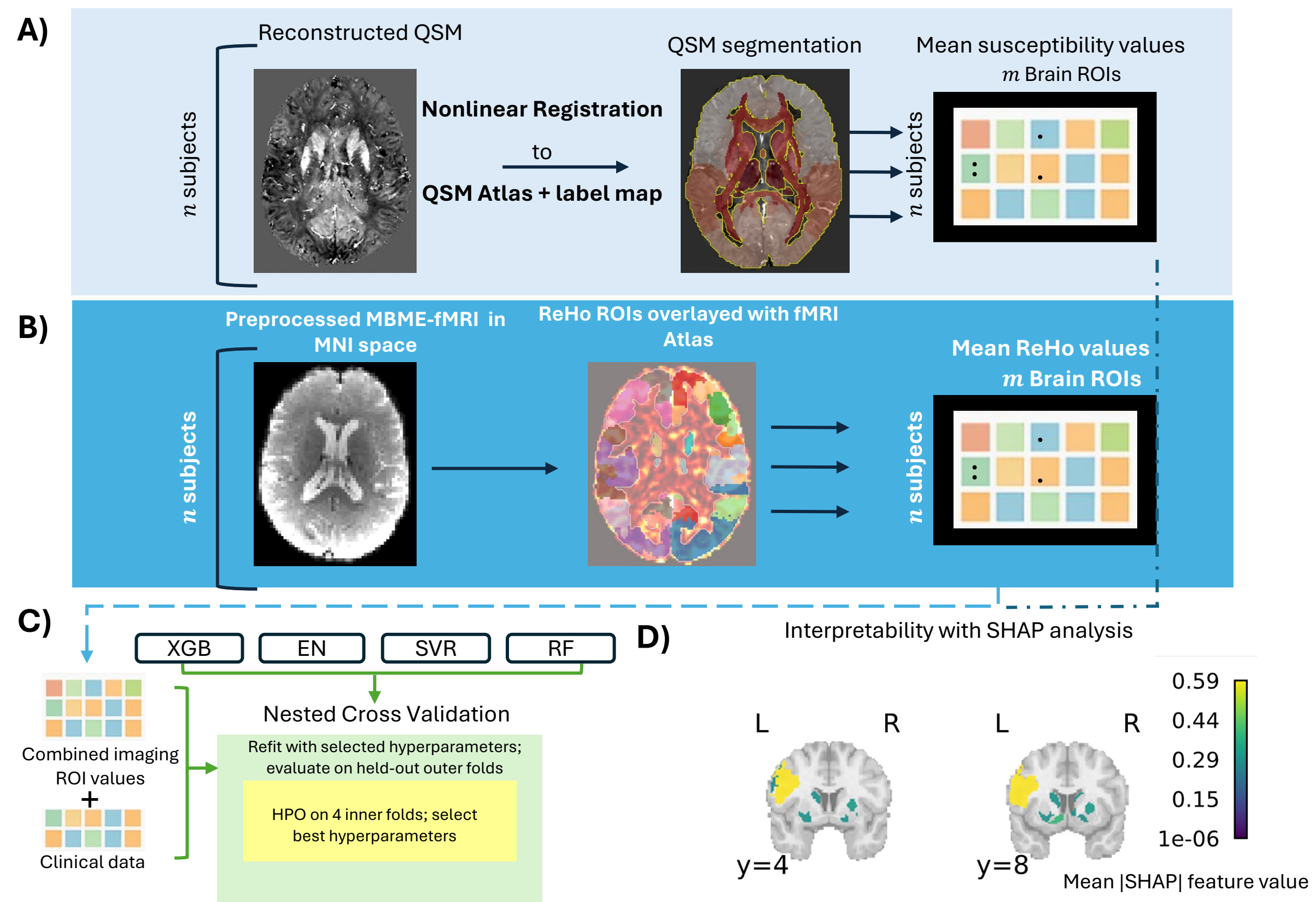

**Figure 1. Overall analysis pipeline.**

**A)** Regional QSM features were extracted by nonlinearly registering each participant's reconstructed QSM image to a QSM atlas and computing mean susceptibility values across atlas-defined brain regions. **B)** Regional ReHo features were derived from preprocessed resting-state MBME fMRI data by computing ReHo maps and extracting mean regional values using the fMRI atlas. **C)** Multimodal feature sets combining QSM, fMRI, and clinical variables were used to predict MDS-UPDRS Part III scores using XGBoost, Elastic Net, Support Vector Regression, and Random Forest models. Models were trained and evaluated using nested cross-validation, with 5 outer folds for held-out performance evaluation and 4 inner folds for hyperparameter optimization. Predictive performance was computed from the held-out outer-fold predictions. **D)** Model interpretation was performed using SHAP analysis to identify features contributing to predicted motor severity.

## Machine leaning pipeline for PD motor severity scores prediction

The overall analysis pipeline is shown in Figure 1. To assess the predictive value of regional QSM, clinical variables, and MBME fMRI, we fit a representative suite of the most powerful classical machine learning models. These were deliberately chosen over deep learning models which tend to have higher data demands. Moreover, Support Vector Regression (SVR)[30] was included to capture potentially nonlinear relationships in a small-sample, high-dimensional setting, Elastic Net provided a regularized linear model with embedded coefficient shrinkage[31], and Random Forest[32] and XGBoost[33] were included as tree-based nonlinear approaches. These selected predictive models were evaluated across 13 combinations of imaging and clinical features (Table S1). The imaging inputs included both full and reduced feature sets: the full sets comprised 206 regional QSM features and 135 regional fMRI features, whereas the reduced sets comprised 89 QSM and 47 fMRI features selected from motor-circuit regions that may be affected by PD, based on either direct evidence from the literature or their known association with motor function (Table S3 and Table S4). The prediction target in all experiments was MDS-UPDRS Part III.

Model performance was evaluated using nested stratified cross-validation with 5 outer folds and 4 inner folds. For each outer fold, predictions were generated for the held-out test participants and then pooled across folds to summarize out-of-sample performance using $R^2$, MAE, RMSE, and Pearson correlation. To complement these conventional regression metrics with a clinically interpretable measure, we also calculated the proportion of participants whose predicted MDS-UPDRS Part III score fell within ±5 points of the observed score. This margin was selected as the CID because prior studies have identified clinically meaningful motor differences on the MDS-UPDRS/UPDRS Part III motor examination in the approximately 4–5-point range[34, 35].

For model interpretation, each selected model was refit on the full dataset using the hyperparameter configuration that achieved the best inner-fold RMSE across outer folds, and SHAP[25] was then used to quantify the contributions of imaging and clinical features to the predicted UPDRS scores. Additional implementation details, including model software, feature scaling, cross-validation split generation and hyperparameter optimization, are provided in Supplementary Methods 1.3. To evaluate whether model performance exceeded chance, we used full-pipeline permutation testing of the outcome labels, as described in Supplementary Methods 1.4.

# Results

Rather than identifying a single universally best model, the results showed that performance depended on the feature set, model class, and evaluation metric. Table S2 reports the extended results across all experiment-model combinations, while Table 2 summarizes the top-performing model for each of the 13 feature sets based on pooled held-out $R^2$. Across these experiments, pooled held-out $R^2$ ranged from 0.023 to 0.454**,** and within-±5-point accuracy ranged from 39.3% to 75.0%.

Permutation testing indicated supported that the observed performance was unlikely to arise by chance. For all top-performing models in Table 2, nested cross-validation permutation testing produced empirical p-values between 0.004975, the resolution for 200 permutations, and 0.049751. Figures S3 (part 1 and 2) show the corresponding permutation null distributions. Multiple-comparison correction was not applied because this was a secondary analysis performed in 13 of the 52 experiment-model combinations, and the limited number of permutations restricted p-value resolution.

Across these experiments, the two definitions of prediction quality were complementary. In Table 2, the strongest global fit was obtained in Expt6, which combined 135 fMRI features, 206 QSM features, and six clinical variables, achieving $R^2 = 0.454$, RMSE = 5.116, MAE = 4.153, $r = 0.675$, and 17/28 predictions within ±5 points (60.7%). In contrast, the highest within-±5 performance was obtained in Expt7, which combined 89 QSM features with clinical variables and achieved $R^2 = 0.374$, RMSE = 5.479, MAE = 3.997, $r = 0.619$, and 21/28 predictions within ±5 points (75.0%). Expt7 also had the lowest MAE among the top-performing models.

**Table 2. Predictive performance of the top-performing model selected for each experiment, with nested cross-validation permutation testing.**

For each experiment, the model with the highest pooled held-out $R^2$ among the four candidate models was selected. Predictive performance is summarized using pooled out-of-fold held-out predictions across the 5 outer folds of nested cross-validation, and reported as pooled held-out $R^2$, RMSE, MAE, and Pearson correlation $r$ between predictions and observed outcomes. Permutation testing was performed for each selected model by rerunning the entire nested cross-validation pipeline, including inner-fold hyperparameter optimization and outer-fold held-out evaluation, under 200 random permutations of the outcome labels across samples while keeping the input features and cross-validation splits fixed. The permutation p-value represents the empirical probability of obtaining a pooled held-out $R^2$ at least as large as the observed value under the shuffled-label null, using the $\frac{n_{extreme}+1}{M+1}$ correction with $M = 200$. $n_{extreme}$ denotes the number of permuted runs whose pooled held-out R2 was greater than or equal to the observed pooled held-out R2. Pearson correlation p-values $p_r$ are shown descriptively. Subjects whose predicted MDS-UPDRS Part III score fell within ±5 points of the observed score of the observed score are reported as the Within ±5 number (n) and proportion (%).

| Expt | Model | Feature set | R2 | RMSE | MAE | r | $p_r$ | Permutation p-value | $n_{extreme}$ | Within ±5 (n) | Within ±5(%) |
|---|---|---|---|---|---|---|---|---|---|---|---|
| Expt1 | SVR | 135 fMRI + Clinical (full) | 0.292 | 5.828 | 4.809 | 0.543 | 0.002829 | 0.004975 | 0 | 14 | 50 |
| Expt2 | Random Forest | 135 fMRI (full) | 0.138 | 6.429 | 5.008 | 0.395 | 0.037358 | 0.009950 | 1 | 16 | 57.143 |
| Expt3 | SVR | 206 QSM + Clinical (full) | 0.211 | 6.149 | 4.605 | 0.46 | 0.013681 | 0.019900 | 3 | 17 | 60.714 |
| Expt4 | Elastic Net | 206 QSM (full) | 0.22 | 6.116 | 4.731 | 0.503 | 0.006332 | 0.014925 | 2 | 15 | 53.571 |
| Expt5 | Elastic Net | Clinical | 0.023 | 6.844 | 5.555 | 0.232 | 0.234256 | 0.049751 | 9 | 11 | 39.286 |
| Expt6 | Elastic Net | 135 fMRI + 206QSM + Clinical (full) | 0.454 | 5.116 | 4.153 | 0.675 | 0.000081 | 0.004975 | 0 | 17 | 60.714 |
| **Expt7** | Elastic Net | Clinical + 89QSM (reduced) | 0.374 | 5.479 | 3.997 | 0.619 | 0.000443 | 0.004975 | 0 | 21 | 75 |
| Expt8 | SVR | 47 fMRI (reduced) | 0.294 | 5.817 | 4.729 | 0.545 | 0.002692 | 0.004975 | 0 | 16 | 57.143 |
| **Expt9** | SVR | 47 fMRI + Clinical (reduced) | 0.405 | 5.34 | 4.135 | 0.638 | 0.000262 | 0.004975 | 0 | 19 | 67.857 |
| Expt10 | SVR | 89 QSM (reduced) | 0.288 | 5.842 | 4.172 | 0.543 | 0.002846 | 0.009950 | 1 | 19 | 67.857 |
| Expt11 | Elastic Net | 89 QSM + 47 fMRI + Clinical (reduced) | 0.402 | 5.354 | 4.417 | 0.642 | 0.000233 | 0.004975 | 0 | 17 | 60.714 |
| Expt12 | Elastic Net | 135 fMRI + 206QSM (full) | 0.423 | 5.26 | 4.331 | 0.654 | 0.000161 | 0.004975 | 0 | 14 | 50 |
| Expt13 | SVR | 89 QSM + 47 fMRI (reduced) | 0.274 | 5.898 | 4.643 | 0.537 | 0.003183 | 0.009950 | 1 | 16 | 57.143 |

Thus, $R^2$ and within-±5 accuracy captured different aspects of model performance. $R^2$ reflected global fit across the cohort, whereas within-±5 accuracy reflected clinically close prediction for individual participants. This distinction is visible in Figure 2 and Supplementary Fig. S1, which show the true versus predicted MDS-UPDRS Part III scores for the top-performing models summarized in Table 2. Figure 2 shows that Expt7 showed the highest proportion of predictions within the ±5-point band, whereas Expt6 and Expt12 showed stronger global fit but lower within-±5 accuracy. Expt12, which had the second-highest $R^2$, also showed shrinkage toward the mean, with lower scores overpredicted and higher scores underpredicted.

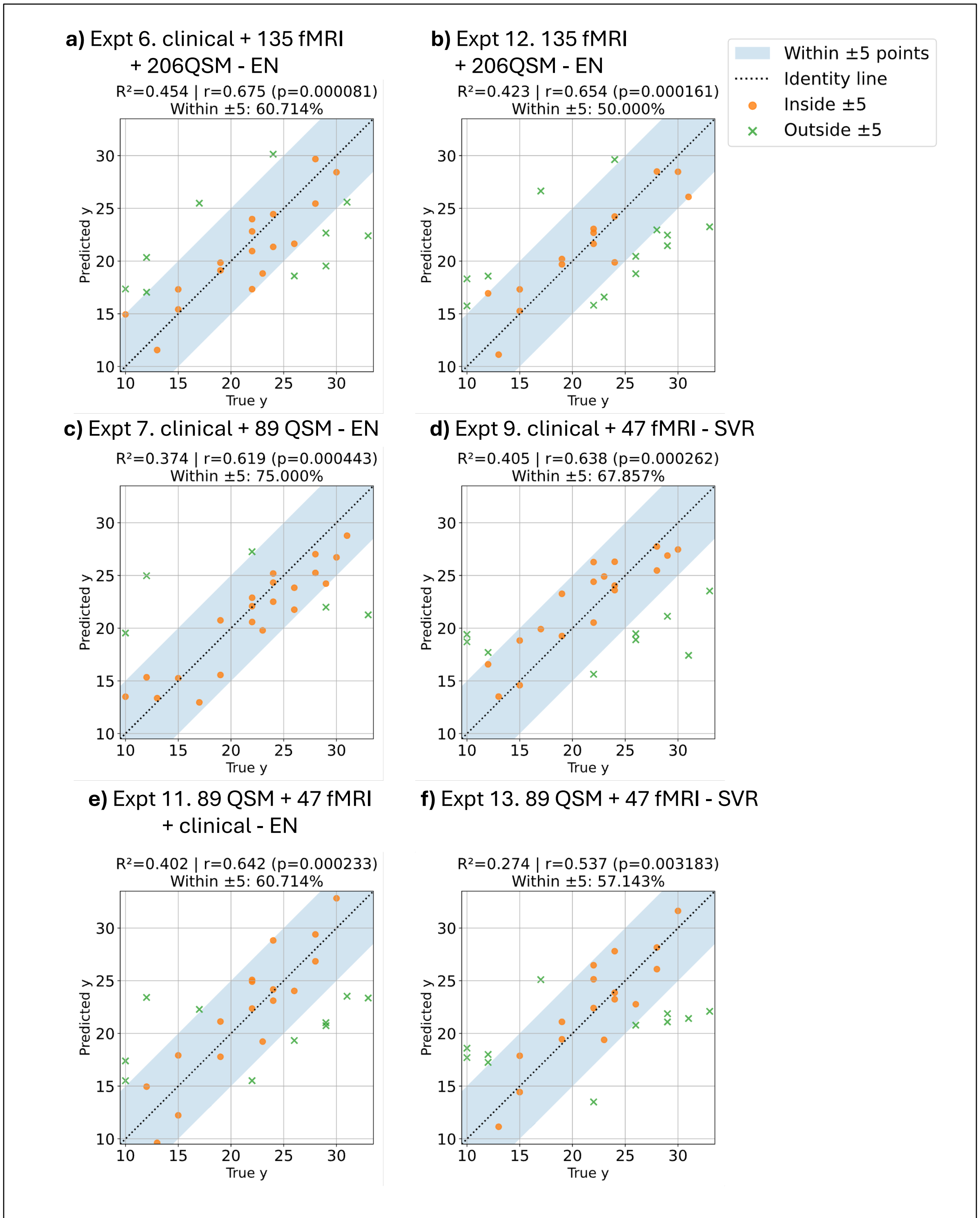

**Figure 2. True versus predicted MDS-UPDRS Part III scores.**

Scores were estimated using nested cross-validation using out-of-fold predictions (5 outer folds; 4 inner folds for hyperparameter optimization). Panels show the top-performing experimental feature-set combinations evaluated for regression of MDS-UPDRS Part III from clinical, fMRI, and QSM data. A shaded blue region marks predictions within ±5 points of the true value, with green points inside and orange points outside this range. Each panel reports held-out $R^2$ ,Pearson correlation ($r$), and corresponding $p$-value. **a)** Expt 6: clinical. **b)** Expt 12: 135fMRI+206QSM **c)** Expt 7: clinical + 89 QSM. **d)** Expt 9: clinical + 47 fMRI. **e)** Expt 11: 89 QSM + 47 fMRI + clinical. **f)** Expt 13: 89QSM+47fMRI. Blue shade corresponds to matching fMRI features and orange shade corresponds to matching QSM features. Green box highlights clinical features.

## Neuroimaging derived features capture disease-relevant motor severity signal

Imaging features alone carried meaningful predictive signal for current motor severity, even without clinical variables. As shown in Table 2, reduced fMRI-derived ReHo features achieved $R^2 = 0.294$ in Expt8, while reduced QSM features achieved $R^2 = 0.288$ in Expt10. Both reduced imaging-only models outperformed their corresponding full single-modality models: full fMRI reached $R^2 = 0.138$ in Expt2, and full QSM reached $R^2 = 0.220$ in Expt4. For QSM, feature reduction also improved clinical closeness, increasing within-±5 performance from 53.6% in Expt4 to 67.9% in Expt10; for fMRI, within-±5 performance remained stable at 57.1% in both Expt2 and Expt8. These results indicate that regional neuroimaging features contained disease-relevant information associated with MDS-UPDRS Part III scores. This signal was not explained by clinical variables alone: the clinical-only model, Expt5, was the weakest single-source model, with $R^2 = 0.023$ and 39.3% of predictions within ±5 points.

## fMRI and QSM provide non-redundant predictive information

The relative performance of fMRI-derived ReHo and QSM depended on the feature set and evaluation metric, indicating non-redundant predictive information rather than clear superiority of one modality. In the full imaging-only comparison, QSM showed stronger global fit than fMRI, with Expt4 reaching $R^2 = 0.220$ compared with $R^2 = 0.138$ in Expt2. However, fMRI had slightly higher clinical closeness, with 57.1% of predictions within ±5 points compared with 53.6% for QSM. In the reduced imaging-only comparison, the two modalities showed similar $R^2$ values, with reduced fMRI reaching $R^2 = 0.294$ in Expt8 and reduced QSM reaching $R^2 = 0.288$ in Expt10. However, reduced QSM produced more clinically close predictions, with 67.9% within ±5 points compared with 57.1% for reduced fMRI. Together, these results show that fMRI and QSM contributed distinct predictive information, consistent with the different functional and susceptibility-related signals measured by each modality.

## Clinical variables add contextual value, especially in reduced imaging models

Clinical variables appeared to refine, rather than replace, imaging-based prediction by adding participant-level context. Their benefit was clearest when combined with reduced, biologically informed imaging feature sets. Selected fMRI improved from $R^2 = 0.294$ and 57.1% within ±5 in Expt8 to $R^2 = 0.405$ and 67.9% within ±5 in Expt9. Similarly, selected QSM improved from $R^2 = 0.288$ and 67.9% within ±5 in Expt10 to $R^2 = 0.374$ and 75.0% within ±5 in Expt7.

Across matched comparisons, adding clinical variables generally improved global fit, including full fMRI, selected multimodal imaging, and full multimodal imaging. However, the effect was not uniform across metrics. For example, adding clinical variables to full QSM slightly decreased $R^2$ from 0.220 in Expt4 to 0.211 in Expt3, although within-±5 performance improved from 53.6% to 60.7%. This pattern indicates that clinical variables improved some models but did not consistently improve clinically close prediction.

## Feature reduction improved single-modality models more consistently than multimodal models

Feature reduction was most beneficial when each imaging modality was modeled separately, supporting the idea that literature-guided feature selection helped reduce noise. For fMRI, reducing the feature set from 135 to 47 regions improved $R^2$ from 0.138 in Expt2 to 0.294 in Expt8, while within-±5 performance remained stable at 57.1%. When clinical variables were included, the reduced fMRI model also outperformed the full fMRI model, as Expt9 reached $R^2 = 0.405$ compared with $R^2 = 0.292$ in Expt1. A similar pattern was observed for QSM: reducing the feature set from 206 to 89 regions improved both $R^2$ and clinical closeness, from $R^2 = 0.220$ and 53.6% within ±5 in Expt4 to $R^2 = 0.288$ and 67.9% within ±5 in Expt10.

In contrast, feature reduction was less clearly beneficial for multimodal models. The full multimodal model with clinical variables, Expt6, achieved the highest overall $R^2 = 0.454$, whereas the selected multimodal model with clinical variables, Expt11, reached $R^2 = 0.402$, although both had the same within-±5 performance of 60.7%. Similarly, full multimodal imaging without clinical variables outperformed reduced multimodal imaging by $R^2$. Together, these findings suggest that feature reduction was most useful for single-modality models, whereas full multimodal integration retained stronger global performance when fMRI and QSM were combined.

## Are multimodal models better than single-modality models?

Multimodal models produced some of the strongest global fits, particularly when imaging features were combined with clinical variables. The full multimodal (fMRI + QSM + clinical model), Expt6, achieved the highest overall $R^2 = 0.454$ in predicting MDS-UPDRS Part III scores. Full multimodal imaging without clinical variables also performed strongly, with Expt12 reaching $R^2 = 0.423$.

However, multimodal models were not consistently superior across all metrics. Selected QSM + clinical variables, Expt7, achieved the highest within-±5 performance despite a lower $R^2$, while reduced multimodal + clinical variables, Expt11, reached a similar $R^2$ to reduced fMRI + clinical variables, Expt9, at 0.402 and 0.405, respectively. Thus, full multimodal integration improved global fit, as shown by Expt6 and Expt12, whereas selected single-modality models, particularly selected QSM + clinical variables in Expt7, produced the most clinically close participant-level predictions.

## Model class performance

In this small-sample, high-dimensional dataset, margin-based SVR and regularized linear Elastic Net models performed more consistently than flexible tree-based methods. As shown in Table 2, SVR was the top-performing model by $R^2$ in 6 of 13 feature sets, Elastic Net in 6 of 13, and Random Forest in 1 of 13. XGBoost was not the top-performing model in any experiment, although it achieved moderate performance in some feature sets, including Expt9.

## Top-ranked SHAP predictors were predominantly imaging-derived

SHAP analyses indicated that the top-ranked predictors were predominantly imaging-derived features. Figure 3 and Supplementary Fig. S2 (part 1 and 2) provide insight into the features underlying these predictions by showing the top 20 most influential features across full and reduced QSM-, fMRI-, and multimodal-based models. Across experiments, several regions appeared repeatedly, including the cerebellum, thalamus, insula, striatum, and other motor-related cortical areas, suggesting that the predictive signal was distributed across a coherent set of motor-relevant and subcortical imaging features.

In the QSM comparison, Expt7 versus Expt10, 15 of the top 20 features overlapped, and LEDD was the only clinical feature appearing among the top-ranked predictors. In the fMRI comparison, Expt9 versus Expt8, the two models also shared several high-ranking imaging features, but Expt9 additionally included LEDD, time since dose, and age. Supplementary Fig. S2 (part 1 and 2) further shows that, across the remaining top-performing models, LEDD and time since dose were the most consistently influential clinical variables when clinical features were included.

# Discussion

In this study, we evaluated whether structural and functional neuroimaging features could predict current Parkinson's disease motor severity, measured by MDS-UPDRS Part III, using QSM, multiband multi-echo fMRI-derived ReHo, and clinical variables. The results support the following main findings:

**First, imaging features alone carried meaningful predictive signal for current MDS-UPDRS Part III scores.** Imaging-only models using reduced fMRI, reduced QSM, and full multimodal imaging explained a meaningful proportion of variance in motor severity, despite the small sample size and high-dimensional feature space. In contrast, the clinical-only model showed weak predictive performance compared with several imaging-only models, suggesting that the strongest predictive signal depended substantially on imaging-derived information.

Prior imaging-based studies predicting Parkinson's disease severity provide useful benchmarks, although direct comparison is limited because the prediction targets differ. Nguyen et al.[16] reported current MDS-UPDRS total score prediction with rs-fMRI ReHo features, achieving $R^2 = 0.304$, while a later replication study by Germani et al. reported lower current-year ReHo performance, with $R^2 = 0.124$[19]. Studies focused on motor severity reported Pearson correlations rather than $R^2$: Hou et al. predicted UPDRS-III from resting-state fMRI with $r = 0.35$[20], and Li et al.[21] predicted post-therapy UPDRS-III with $r = 0.65$ in the medication-on condition, with MAE = 7.52. In comparison, our full multimodal fMRI, QSM, and clinical model predicted current MDS-UPDRS Part III scores with $R^2 = 0.454$, $r = 0.675$, and MAE = 4.153, suggesting competitive performance while extending prior work by integrating motion-corrected QSM with multiband multi-echo fMRI-derived ReHo.

**Second, fMRI and QSM appeared non-redundant rather than clearly superior across all metrics.** The full multimodal model with clinical variables achieved the strongest global fit, whereas selected QSM plus clinical variables produced the highest within-±5-point accuracy and lowest MAE among the top-performing models. This distinction suggests that $R^2$ and clinically close prediction capture different aspects of model performance and should be interpreted together. Biologically, this non-redundancy is plausible because QSM captures susceptibility-related structural variation, including iron-sensitive changes in PD-relevant regions, whereas fMRI-derived ReHo reflects local synchronization of spontaneous BOLD signal fluctuations.

**Third, clinical variables provided useful context, but they were not sufficient on their own.** Third, clinical variables provided useful context but were not sufficient on their own. Adding them improved performance in most matched imaging comparisons, particularly for reduced fMRI and reduced QSM feature sets, suggesting that clinical information helped contextualize imaging-derived signal. This effect was especially evident for fMRI-based models, likely because functional activity may be influenced by medication state and other participant-level factors, whereas QSM provides a more stable, iron-sensitive measure of structural tissue properties. Thus, clinical variables may contribute complementary information related to levodopa dosage, time since medication intake, and current motor function in Parkinson's disease.

**Fourth, biologically informed feature reduction improved single-modality prediction.** In high-dimensional neuroimaging data, reducing the feature space can limit overfitting and improve generalization, particularly in lower-N settings. This was evident for both fMRI and QSM. For fMRI, reducing the feature set from 135 to 47 regions improved performance in the imaging-only comparison, and the benefit was even clearer when clinical variables were included. For QSM, reducing the feature set from 206 to 89 regions improved both $R^2$ and within-±5-point accuracy.

**Fifth, SHAP analysis supported the biological plausibility of the predictive models.** Across experiments, top-ranked predictors often overlapped across related models and remained strongly enriched for imaging features even when clinical variables were included. Influential regions repeatedly included the cerebellum, striatum, insula, and thalamic regions, all of which are known to be involved in Parkinson's disease motor dysfunction [36-39].The repeated appearance of these regions suggests that the models captured disease-relevant neuroimaging variation rather than relying on unstable or isolated predictors. The SHAP results also showed that clinical variables, including LEDD, time since dose, and age, contributed in some models, but did not dominate the feature rankings.

This study has several limitations. The small sample size increased the risk of overfitting and limited the stability and generalizability of model estimates, although nested cross-validation and permutation testing were used to reduce bias and assess chance-level performance. In addition, image registration and alignment could be further improved to increase the precision of regional feature extraction. We also tested only one atlas per imaging contrast, and other parcellation strategies may capture disease-relevant variation differently. Finally, we focused on classical machine learning models because the cohort size was not well suited for deep learning or foundation model fine-tuning. Future studies with larger, multisite cohorts, external validation, alternative atlases, and additional model classes will be needed to confirm the robustness of these findings.

This work supports the use of QSM and multiband multi-echo fMRI-derived ReHo as non-redundant imaging biomarkers for estimating current Parkinson's disease motor severity. Full multimodal imaging plus clinical variables provided the strongest global fit, while selected QSM plus clinical variables produced the most clinically close predictions. Across models, imaging-derived features remained central to prediction, clinical variables provided useful contextual information, and SHAP analyses identified biologically plausible regions involved in Parkinson's disease motor dysfunction. These findings support a framework in which structural and functional neuroimaging features are used not only to predict motor severity, but also to identify interpretable brain regions associated with individual differences in Parkinson's disease motor impairment.

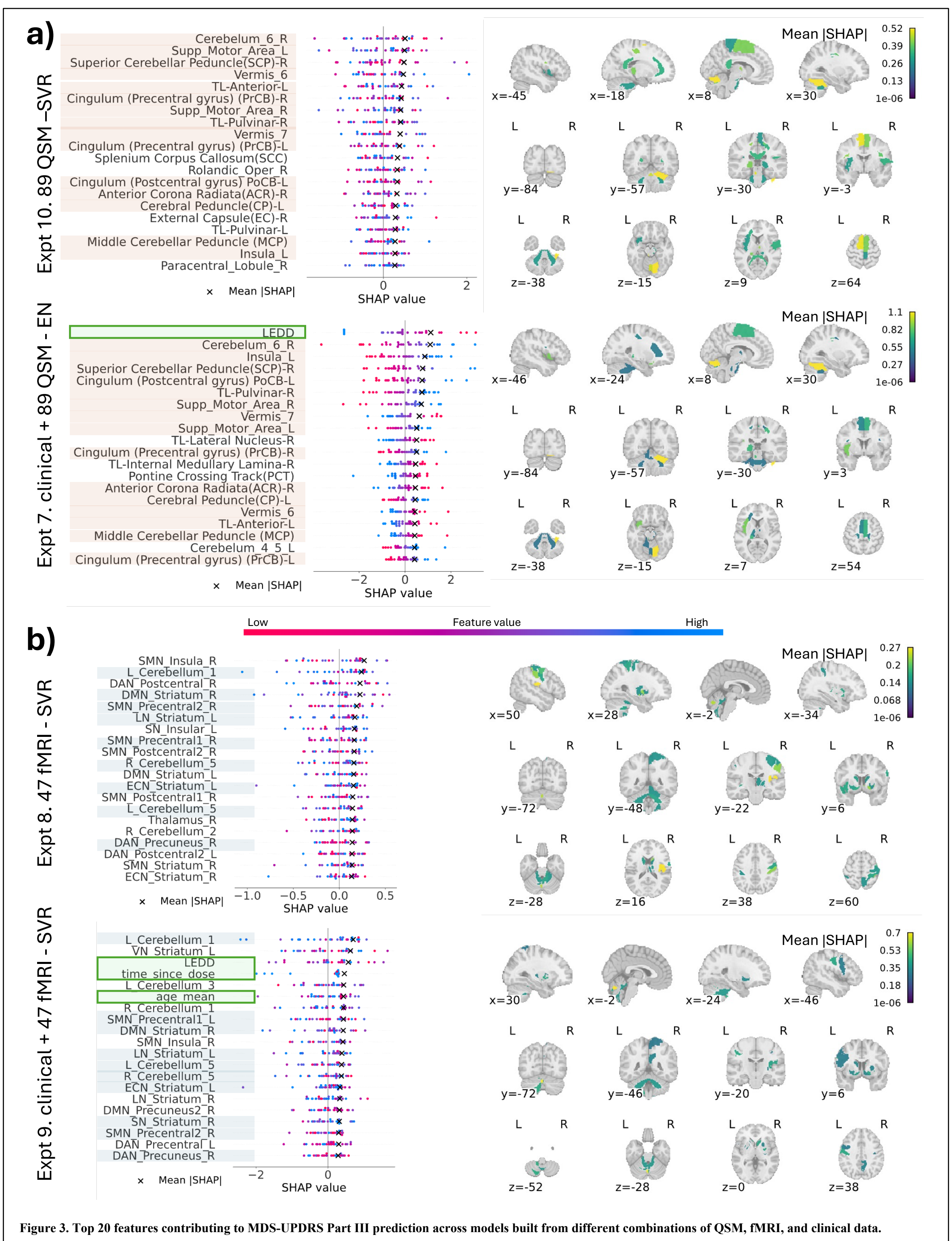

**Figure 3. Top 20 features contributing to MDS-UPDRS Part III prediction across models built from different combinations of QSM, fMRI, and clinical data.**

Left subpanels show per-feature SHAP value distributions across subjects, with color indicating feature magnitude (red = high, blue = low) and black X denoting mean absolute SHAP. Each point represents one subject's SHAP value for that feature; positive SHAP values indicate contributions toward higher predicted MDS-UPDRS Part III scores, whereas negative values indicate contributions toward lower predicted scores. Right subpanels show the spatial distribution of mean absolute SHAP values projected onto MNI brain slices for the corresponding imaging features. **a)** Expt 7: clinical + 89 QSM and Expt 10: 89 QSM. **b)** Expt 8: 47 fMRI and Expt 9: clinical + 47 fMRI.

## Acknowledgments

The authors acknowledge the University of Texas Southwestern BioHPC High Performance Computing facility for providing computational resources and technical support used in this study. We also thank the clinical and imaging teams involved in participant assessment, data collection, and study coordination.

## Ethics Statement

The study protocol was approved by the Institutional Review Board of University of Texas Medical Center under protocol STU2018-0428. All imaging and clinical data were collected and analyzed in accordance with institutional guidelines and the principles of the Declaration of Helsinki. Written informed consent was obtained from all participants before study participation.

# References

1. Kalia, L.V. and A.E. Lang, *Parkinson's disease.* Lancet, 2015. **386**(9996): p. 896-912.
2. Goetz, C.G., et al., *Movement Disorder Society-sponsored revision of the Unified Parkinson's Disease Rating Scale (MDS-UPDRS): scale presentation and clinimetric testing results.* Mov Disord, 2008. **23**(15): p. 2129-70.
3. Madelung, C.F., et al., *High-resolution mapping of substantia nigra in Parkinson's disease using 7 tesla magnetic resonance imaging.* NPJ Parkinsons Dis, 2025. **11**(1): p. 113.
4. Riederer, P., et al., *Lewy bodies, iron, inflammation and neuromelanin: pathological aspects underlying Parkinson's disease.* J Neural Transm (Vienna), 2023. **130**(5): p. 627-646.
5. Schweser, F., et al., *Quantitative susceptibility mapping for investigating subtle susceptibility variations in the human brain.* Neuroimage, 2012. **62**(3): p. 2083-100.
6. Guo, W., et al., *Quantitative susceptibility mapping of subcortical iron deposition in Parkinson disease and multiple system atrophy: clinical correlations and diagnostic implications.* Quant Imaging Med Surg, 2024. **14**(7): p. 4464-4474.
7. Zhao, H., et al., *Association between deep gray matter iron deposition and clinical symptoms in Parkinson's disease: a quantitative susceptibility mapping study.* Front Neurol, 2024. **15**: p. 1442903.
8. Risk, B.B., et al., *Which multiband factor should you choose for your resting-state fMRI study?* Neuroimage, 2021. **234**: p. 117965.
9. Kundu, P., et al., *Multi-echo fMRI: A review of applications in fMRI denoising and analysis of BOLD signals.* Neuroimage, 2017. **154**: p. 59-80.
10. DuPre, E., et al., *TE-dependent analysis of multi-echo fMRI with tedana.* Journal of Open Source Software, 2021. **6**(66).
11. Zang, Y., et al., *Regional homogeneity approach to fMRI data analysis.* Neuroimage, 2004. **22**(1): p. 394-400.
12. Yuan, L.-X., et al. *Reliability and validity of multi-band multi-echo fMRI*. bioRxiv, 2026. DOI: 10.64898/2026.01.16.699838.
13. Mirelman, A., et al., *Digital Mobility Measures: A Window into Real-World Severity and Progression of Parkinson's Disease.* Mov Disord, 2024. **39**(2): p. 328-338.

14. ZhuParris, A., et al., *Treatment Detection and Movement Disorder Society-Unified Parkinson's Disease Rating Scale, Part III Estimation Using Finger Tapping Tasks.* Mov Disord, 2023. **38**(10): p. 1795-1805.
15. Mirelman, A., et al., *Detecting Sensitive Mobility Features for Parkinson's Disease Stages Via Machine Learning.* Mov Disord, 2021. **36**(9): p. 2144-2155.
16. Nguyen, K.P., et al., *Predicting Parkinson's disease trajectory using clinical and neuroimaging baseline measures.* Parkinsonism Relat Disord, 2021. **85**: p. 44-51.
17. Loo, R.T.J., et al., *Interpretable Machine Learning for Cross-Cohort Prediction of Motor Fluctuations in Parkinson's Disease.* Mov Disord, 2025. **40**(8): p. 1604-1617.
18. Gajos, K.Z., et al., *Computer mouse use captures ataxia and parkinsonism, enabling accurate measurement and detection.* Mov Disord, 2020. **35**(2): p. 354-358.
19. Germani, E., et al., *Predicting Parkinson's disease trajectory using clinical and functional MRI features: A reproduction and replication study.* PLoS One, 2025. **20**(2): p. e0317566.
20. Hou, Y., et al., *Prediction of individual clinical scores in patients with Parkinson's disease using resting-state functional magnetic resonance imaging.* J Neurol Sci, 2016. **366**: p. 27-32.
21. Li, X., et al., *Predicting the Post-therapy Severity Level (UPDRS-III) of Patients With Parkinson's Disease After Drug Therapy by Using the Dynamic Connectivity Efficiency of fMRI.* Front Neurol, 2019. **10**: p. 668.
22. Ragothaman, A., et al., *Motor networks, but also non-motor networks predict motor signs in Parkinson's disease.* Neuroimage Clin, 2023. **40**: p. 103541.
23. Mellema, C.J., et al., *Longitudinal prognosis of Parkinson's outcomes using causal connectivity.* Neuroimage Clin, 2024. **42**: p. 103571.
24. Abbasi, N., et al., *Predicting severity and prognosis in Parkinson's disease from brain microstructure and connectivity.* Neuroimage Clin, 2020. **25**: p. 102111.
25. Lundberg, S. and S. Lee *A Unified Approach to Interpreting Model Predictions*. 2017.
26. Mao, Z., et al., *Population pharmacodynamics of IPX066: an oral extended-release capsule formulation of carbidopa-levodopa, and immediate-release carbidopa-levodopa in patients with advanced Parkinson's disease.* J Clin Pharmacol, 2013. **53**(5): p. 523-31.
27. Cilia, R., et al., *Natural history of motor symptoms in Parkinson's disease and the long-duration response to levodopa.* Brain, 2020. **143**(8): p. 2490-2501.
28. Zhang, Y., et al., *Longitudinal atlas for normative human brain development and aging over the lifespan using quantitative susceptibility mapping.* Neuroimage, 2018. **171**: p. 176-189.
29. Schaefer, A., et al., *Local-Global Parcellation of the Human Cerebral Cortex from Intrinsic Functional Connectivity MRI.* Cereb Cortex, 2018. **28**(9): p. 3095-3114.
30. Cortes, C. and V. Vapnik, *Support-vector networks.* Machine Learning, 1995. **20**(3): p. 273-297.
31. Zou, H. and T. Hastie, *Regularization and Variable Selection Via the Elastic Net.* Journal of the Royal Statistical Society Series B: Statistical Methodology, 2005. **67**(2): p. 301-320.
32. Breiman, L., *Random Forests.* Machine Learning, 2001. **45**(1): p. 5-32.
33. Chen, T. and C. Guestrin. *Xgboost: a scalable tree boosting system*. 2016.
34. Sanchez-Ferro, A., et al., *Minimal Clinically Important Difference for UPDRS-III in Daily Practice.* Mov Disord Clin Pract, 2018. **5**(4): p. 448-450.
35. Trundell, D., et al., *Estimation of and clinical consensus on the meaningful motor progression threshold on MDS-UPDRS Part III.* J Parkinsons Dis, 2025. **15**(1): p. 97-110.

36. Wu, T. and M. Hallett, *The cerebellum in Parkinson's disease.* Brain, 2013. **136**(Pt 3): p. 696-709.
37. Criaud, M., et al., *Contribution of insula in Parkinson's disease: A quantitative meta-analysis study.* Hum Brain Mapp, 2016. **37**(4): p. 1375-92.
38. Zhai, S., et al., *Striatal synapses, circuits, and Parkinson's disease.* Curr Opin Neurobiol, 2018. **48**: p. 9-16.
39. Bosch-Bouju, C., B.I. Hyland, and L.C. Parr-Brownlie, *Motor thalamus integration of cortical, cerebellar and basal ganglia information: implications for normal and parkinsonian conditions.* Front Comput Neurosci, 2013. **7**: p. 163.

# Supplementary Information

## INTERPRETABLE MACHINE LEARNING PREDICTS PARKINSON'S DISEASE SEVERITY USING MOTION-CORRECTED QSM MRI AND MULTIBAND MULTI-ECHO FMRI FEATURES

*Aixa X. Andrade*[1,2,3]

AFFILIATIONS
[1]Lyda Hill Department of Bioinformatics
[2]Department of Biomedical Engineering
[3]University of Texas Southwestern Medical Center, Dallas, Texas, USA

# Table of Contents

# 1. Supplementary Methods

## 1.1. Imaging and preprocessing QSM

A tuned multishot multi-echo 3D Eco Planar Imaging (EPI) T2*-weighted MRI pulse sequence (Huang, Chen et al. 2024) was acquired on a Siemens MAGNETOM Prisma scanner and used to retrospectively ameliorate artifacts caused by motion and magnetic field fluctuations in the T2*-weighted data and subsequent QSM image reconstruction was performed using the method detailed in Huang et al., 2024(Huang, Chen et al. 2024) using core motion correction principles from van Gelderen et al., 2023(van Gelderen, Li et al. 2023). Figure 3.1A illustrates high intranuclear contrast (e.g., substantia nigra, globus pallidus, red nucleus and parts of the thalamus)(Guo, Zhang et al. 2024) that can be achieved by such advanced QSM reconstruction. This is evident in PD-relevant regions—particularly within the basal ganglia—which is valuable given prior reports of QSM's sensitivity to iron accumulation associated with dopaminergic neurodegeneration(Riederer, Nagatsu et al. 2023).

For registration, QSM data was rotated to LPI orientation, and a multistage normalization pipeline was applied that included N4 bias field correction to correct for low-frequency intensity nonuniformities, followed by histogram matching to improve intensity correspondence. The images were nonlinearly registered to a QSM atlas in MNI space(Zhang, Wei et al. 2018) using the Symmetric Normalization (SyN) implemented in ANTSPy(Avants, Tustison et al. 2011).

Regions of interest (ROIs) were defined from a standard QSM atlas which included 206 bilateral structures (Zhang, Wei et al. 2018). This atlas label map was inverse-warped to each participant's native QSM space, and mean magnetic susceptibility was extracted per ROI using Nilearn's NiftiLabelsMasker(Abraham, Pedregosa et al. 2014).

## 1.2. Imaging and preprocessing Multiband Multiecho fMRI

Resting-state functional MRI data were acquired on a Siemens MAGNETOM Prisma scanner using a multi-band multi-echo EPI BOLD sequence. The resting-state acquisition used 56 interleaved slices with 2.4 mm isotropic resolution, repetition time TR = 1598 ms, flip angle = 60°, multi-band acceleration factor = 4, and anterior-to-posterior phase encoding (AP; from the front of the head toward the back). Four echoes were acquired at echo time TE = 16.80, 39.94, 63.08, and 86.22 ms, with 244 measurements collected during the resting-state run. The acquisition used a 216 mm field of view, base resolution of 90, GeneRalized Autocalibrating Partially Parallel Acquisition (GRAPPA) acceleration factor 2, and magnitude reconstruction. Opposite phase-encoding EPI field-map images were also acquired in AP and PA directions with matched spatial resolution and echo times.

Resting-state multi-echo fMRI data were preprocessed using a Nipype-based workflow(Gorgolewski, Burns et al. 2011). Slice-timing correction was performed for each echo. Inter-volume head motion was corrected by estimating rigid-body realignment parameters from the first echo using MCFLIRT(Jenkinson, Bannister et al. 2002), and the resulting transformations were then applied to all echoes. Functional skull stripping was performed on the mean reference echo by generating a native-space brain mask from the intersection of masks produced by FSL BET(Smith 2002) and AFNI 3dAutomask(Cox 1996); the resulting mask was then applied to the realigned functional data. The skull-stripped mean reference echo was nonlinearly spatially normalized to a reference asymmetric nonlinear MNI template (Fonov, Evans et al. 2009) using ANTs-based EPI normalization(Avants, Tustison et al. 2011). Multi-echo denoising and optimal echo combination were then performed with tedana(DuPre, Salo et al. 2021), followed by minimum image regression to further reduce noise. The tedana-denoised image was transformed to MNI space using the EPI normalization transform and smoothed with a 3.0-mm full width half max (FWHM) Gaussian kernel.

The preprocessed MBME fMRI was bandpass filtered between 0.009 and 0.08 Hz. C-PAC(Craddock, Sikka et al. 2013) was used to compute ReHo(Zang, Jiang et al. 2004) with 27 nearest neighbors. Mean regional BOLD time series were extracted using the 135 Schaefer 2018 atlas with subcortical labels(Schaefer, Kong et al. 2018, Nguyen, Raval et al. 2021). This atlas was chosen because it provides a network-based brain parcellation suitable for summarizing regional resting-state fMRI measures. Regional values were extracted with NiftiLabelsMasker(Abraham, Pedregosa et al. 2014), yielding one mean ReHo value per region for each subject.

## 1.3. Machine learning implementation, cross-validation, and hyperparameter optimization

Support Vector Regression (SVR), Elastic Net, and Random Forest were implemented with scikit-learn(Pedregosa, Varoquaux et al. 2012), while the eXtreme Gradient Boosting (XGBoost) model was implemented with the xgboost library(Chen and Guestrin 2016) and accesed via its scikit-learn-compatible XGBRegressor interface. SVR was included to capture potentially nonlinear relationships in a small-sample, high-dimensional setting, Elastic Net provided a regularized linear model with embedded coefficient shrinkage, and Random Forest and XGBoost were included as tree-based nonlinear approaches. Model training and evaluation were performed on a pre-scaled feature matrix using nested stratified cross-validation with 5 outer folds and 4 inner folds. Nested cross validation was chosen because it yields one of the least biased estimates of real-world performance(Varma and Simon 2006). The nested splits were generated once using stratification labels derived from Diagnosis, sex, and quantile-binned age, and were then saved and reloaded for downstream analyses. Feature scaling was performed with scikit-learn StandardScaler on the numeric columns of the feature matrix, whereas dummy-coded categorical variables were left unscaled. For each outer training split, hyperparameters were optimized in the inner loop using Optuna(Akiba, Sano et al. 2019) with 100 trials (hyperparameter configuration) per predictive model class and RMSE as the tuning objective. The hyperparameter configuration with the lowest mean inner-fold RMSE was then used to refit the model on the corresponding outer-training split, and predictions were generated for the held-out outer-test split. Predictive performance was summarized from the pooled out-of-fold predictions across the outer folds using the coefficient of determination ($R^2$), mean absolute error (MAE), root mean square error (RMSE), and Pearson correlation coefficient (r) with its associated p value.

## 1.4. Permutation testing

To rigorously evaluate whether any predictive model learned the association by mere chance, we performed permutation testing by rerunning the entire nested cross-validation and hyperparameter optimization pipeline after shuffling the outcome labels across subjects. In each permutation run, the feature matrix and predefined fold assignments were kept fixed, but the correspondence between each subject's features and outcome was randomly permuted. The permuted outcome vector was then used throughout the full analysis for that run, including inner-loop hyperparameter tuning, outer-loop model fitting, and held-out outer-fold evaluation. This yielded a null distribution of pooled out-of-fold $R^2$ values, which was compared with the observed pooled out-of-fold $R^2$ from the original, unpermuted analysis. The empirical permutation p value was computed as $p = (n_{\text{extreme}} + 1)/(M + 1)$, where $n_{\text{extreme}}$is the number of permuted runs with pooled out-of-fold $R^2$ greater than or equal to the observed value and $M = 200$ is the total number of permutations. Because of this setup, the p-value resolution is $\frac{1}{201} = 0.005$.

# 2. Supplementary Results

## 2.4. Supplementary Tables

**Table S1. Experimental feature-set combinations used for Parkinson's disease motor severity prediction.**

| Experiment | Feature set | Purpose |
|---|---|---|
| Expt 1 | 135 fMRI + clinical | Tests whether full regional fMRI features improve prediction when combined with clinical variables. |
| Expt 2 | 135 fMRI alone | Evaluates the predictive signal of the full fMRI feature set alone. |
| Expt 3 | 206 QSM + clinical | Tests whether full regional QSM features improve prediction when combined with clinical variables. |
| Expt 4 | 206 QSM alone | Evaluates the predictive signal of the full QSM feature set alone. |
| Expt 5 | Clinical alone | Provides a clinical-only ablation model for comparison with imaging-based models. |
| Expt 6 | 135 fMRI + 206 QSM + clinical | Evaluates the combined contribution of full fMRI, full QSM, and clinical variables. |
| Expt 7 | Clinical + 89 QSM | Tests whether a reduced, biologically informed QSM feature set improves prediction with clinical variables. |
| Expt 8 | 47 fMRI alone | Evaluates the predictive signal of the reduced, biologically informed fMRI feature set alone. |
| Expt 9 | 47 fMRI + clinical | Tests whether a reduced, biologically informed fMRI feature set improves prediction with clinical variables. |
| Expt 10 | 89 QSM alone | Evaluates the predictive signal of the reduced, biologically informed QSM feature set alone. |
| Expt 11 | 89 QSM + 47 fMRI + clinical | Evaluates the combined contribution of reduced QSM, reduced fMRI, and clinical variables. |
| Expt 12 | 135 fMRI + 206 QSM | Tests the combined predictive signal of full fMRI and full QSM features. |
| Expt 13 | 89 QSM + 47 fMRI | Tests the combined predictive signal of reduced QSM and reduced fMRI features. |

**Table S2. Comparison of predictive performance across all 52 experiment–model combinations.**

Predictive performance is summarized using pooled out-of-fold held-out predictions across the 5 outer folds of nested cross-validation, and reported as pooled held-out $R^2$, RMSE, MAE, and Pearson correlation (r) and associated p value $p_r$ between predictions and observed outcomes.

| Expt | Model | R2 | RMSE | MAE | r | p_r |
|---|---|---|---|---|---|---|
| Expt1B | **SVR** | **0.292** | 5.828 | 4.809 | 0.543 | 0.002829 |
| Expt1B | ElasticNet | 0.102 | 6.562 | 5.428 | 0.356 | 0.062669 |
| Expt1B | Random Forest | 0.163 | 6.336 | 4.93 | 0.416 | 0.027668 |
| Expt1B | XGBoost | 0.09 | 6.607 | 5.386 | 0.373 | 0.050625 |
| Expt2 | SVR | 0.072 | 6.67 | 5.636 | 0.34 | 0.077037 |
| Expt2 | ElasticNet | -0.065 | 7.145 | 6.028 | 0.126 | 0.523290 |
| Expt2 | **Random Forest** | **0.138** | 6.429 | 5.008 | 0.395 | 0.037358 |
| Expt2 | XGBoost | -0.045 | 7.079 | 5.693 | 0.261 | 0.180087 |
| Expt3B | **SVR** | **0.211** | 6.149 | 4.605 | 0.46 | 0.013681 |
| Expt3B | ElasticNet | -0.037 | 7.052 | 5.539 | 0.35 | 0.067942 |
| Expt3B | Random Forest | -0.045 | 7.078 | 5.719 | 0.039 | 0.844354 |
| Expt3B | XGBoost | 0.032 | 6.815 | 5.504 | 0.282 | 0.145987 |
| Expt4 | SVR | 0.18 | 6.27 | 4.898 | 0.426 | 0.023721 |
| Expt4 | **ElasticNet** | **0.22** | 6.116 | 4.731 | 0.503 | 0.006332 |
| Expt4 | Random Forest | -0.1 | 7.261 | 5.605 | 0.048 | 0.807852 |
| Expt4 | XGBoost | 0.015 | 6.874 | 5.322 | 0.283 | 0.144221 |
| Expt5B | SVR | -0.071 | 7.167 | 5.954 | 0.178 | 0.364772 |
| Expt5B | **ElasticNet** | **0.023** | 6.844 | 5.555 | 0.232 | 0.234256 |
| Expt5B | Random Forest | -0.148 | 7.42 | 6.386 | 0.05 | 0.800994 |
| Expt5B | XGBoost | -0.322 | 7.962 | 6.657 | -0.05 | 0.800945 |
| Expt6B | SVR | 0.358 | 5.55 | 4.449 | 0.601 | 0.000722 |
| Expt6B | **ElasticNet** | **0.454** | 5.116 | 4.153 | 0.675 | 0.000081 |
| Expt6B | Random Forest | 0.074 | 6.662 | 5.267 | 0.295 | 0.127655 |
| Expt6B | XGBoost | -0.014 | 6.975 | 5.657 | 0.238 | 0.222838 |
| Expt7B | SVR | 0.224 | 6.102 | 4.62 | 0.505 | 0.006107 |
| Expt7B | **ElasticNet** | **0.374** | 5.479 | 3.997 | 0.619 | 0.000443 |
| Expt7B | Random Forest | 0 | 6.924 | 5.473 | 0.161 | 0.414095 |
| Expt7B | XGBoost | 0.003 | 6.915 | 5.309 | 0.28 | 0.148960 |
| Expt8 | **SVR** | **0.294** | 5.817 | 4.729 | 0.545 | 0.002692 |
| Expt8 | ElasticNet | 0.118 | 6.502 | 5.43 | 0.365 | 0.055992 |
| Expt8 | Random Forest | 0.125 | 6.478 | 5.218 | 0.412 | 0.029517 |
| Expt8 | XGBoost | 0.211 | 6.152 | 4.96 | 0.496 | 0.007295 |
| Expt9B | **SVR** | **0.405** | 5.34 | 4.135 | 0.638 | 0.000262 |
| Expt9B | ElasticNet | 0.238 | 6.047 | 4.922 | 0.49 | 0.008132 |
| Expt9B | Random Forest | 0.232 | 6.069 | 4.796 | 0.483 | 0.009177 |
| Expt9B | XGBoost | 0.307 | 5.767 | 4.488 | 0.56 | 0.001932 |
| Expt10 | **SVR** | **0.288** | 5.842 | 4.172 | 0.543 | 0.002846 |
| Expt10 | ElasticNet | 0.241 | 6.034 | 4.53 | 0.528 | 0.003872 |
| Expt10 | Random Forest | -0.009 | 6.955 | 5.223 | 0.197 | 0.315135 |
| Expt10 | XGBoost | -0.055 | 7.113 | 5.062 | 0.238 | 0.222727 |
| Expt11B | SVR | 0.366 | 5.516 | 4.528 | 0.611 | 0.000560 |
| Expt11B | **ElasticNet** | **0.402** | 5.354 | 4.417 | 0.642 | 0.000233 |
| Expt11B | Random Forest | 0.155 | 6.364 | 5.011 | 0.394 | 0.037786 |
| Expt11B | XGBoost | 0.127 | 6.47 | 4.916 | 0.404 | 0.032879 |
| Expt12 | SVR | 0.32 | 5.712 | 4.652 | 0.566 | 0.001684 |
| Expt12 | **ElasticNet** | **0.423** | 5.26 | 4.331 | 0.654 | 0.000161 |
| Expt12 | Random Forest | 0.1 | 6.568 | 5.218 | 0.335 | 0.081498 |
| Expt12 | XGBoost | 0.094 | 6.592 | 5.449 | 0.356 | 0.062712 |
| Expt13 | **SVR** | **0.274** | 5.898 | 4.643 | 0.537 | 0.003183 |
| Expt13 | ElasticNet | 0.21 | 6.154 | 4.936 | 0.511 | 0.005508 |
| Expt13 | Random Forest | 0.041 | 6.781 | 5.293 | 0.284 | 0.142773 |
| Expt13 | XGBoost | 0.118 | 6.502 | 4.896 | 0.395 | 0.037734 |

**Table S3. Condensed summary of the 89 quantitative susceptibility mapping atlas features selected for their relevance to motor dysfunction in Parkinson's disease.**

A total of 89 Parkinson's disease- and motor-relevant regions of interest were selected from the 206 labeled regions in the Zhang et al. 2018 atlas(Zhang, Wei et al. 2018) based on their involvement in motor, sensorimotor, cerebellar, basal ganglia, thalamo-cortical, and white matter pathways that may be affected in Parkinson's disease. For simplicity, bilateral and anatomically related ROIs are presented as grouped entries, while the selected feature set comprises 89 individual atlas-derived features. Left and right labels indicate hemisphere.

| ROI(s) | Impact on the motor circuit |
|---|---|
| SNpc-L/R | Basal ganglia motor circuit(McGregor and Nelson 2019) |
| SNpr-L/R | Basal ganglia motor circuit(Galvan, Devergnas et al. 2015) |
| Putamen-L/R | Basal ganglia motor circuit(Galvan, Devergnas et al. 2015) |
| Globus pallidus-L/R | Basal ganglia motor circuit(Galvan, Devergnas et al. 2015) |
| Caudate-L/R | Basal ganglia motor planning and control(Grahn, Parkinson et al. 2008) |
| Red nucleus-L/R | Midbrain and cerebellar motor coordination(Basile, Quartu et al. 2021) |
| Thalamic nuclei: anterior, medial, midline, pulvinar, internal medullary lamina, lateral nucleus L/R | Thalamo-cortical / basal ganglia–cerebellar integration(Bosch-Bouju, Hyland et al. 2013) |
| Precentral gyrus L/R | Primary motor cortex(Chu, Smith et al. 2024) |
| Supplementary motor area L/R | Motor planning circuit(Rahimpour, Rajkumar et al. 2022) |
| Paracentral lobule L/R | Sensorimotor cortex / lower-limb motor control(Kimura, Yamada et al. 2024) |
| Postcentral gyrus L/R | Somatosensory cortex / sensorimotor integration(Borich, Brodie et al. 2015) |
| Rolandic operculum L/R | Sensorimotor integration region(Maliia, Donos et al. 2018, Zhang, Li et al. 2022) |
| Middle cingulate L/R | Motor control(Hoffstaedter, Grefkes et al. 2014) |
| Insula L/R | Sensorimotor integration(Criaud, Christopher et al. 2016, Tinaz, Para et al. 2018) |
| Dentate nucleus L/R | Cerebellar output signals(He, Huang et al. 2017) |
| Superior cerebellar peduncle (SCP) L/R | Cerebellar output pathway(Pietracupa, Ojha et al. 2024) |
| Middle cerebellar peduncle (MCP) | Cortico-ponto-cerebellar motor circuit (input)(Tamanini, Ribeiro et al. 2024) |
| Inferior cerebellar peduncle (ICP) L/R | Cerebellar input connectivity(Jang and Lee 2020) |
| Pontine crossing tract | Cortico-ponto-cerebellar motor circuit(Seo and Jang 2013) |
| Cerebellum 4-5, 6, 8 L/R | Cerebellum(Wu and Hallett 2013, Guell, Schmahmann et al. 2018) |
| Vermis 4-5,6 ,7,8 | Axial/postural cerebellar control(Park, Lee et al. 2018) |
| Corticospinal tract (CST) L/R | Descending motor pathway(Xu, Ding et al. 2020) |
| Cerebral peduncle L/R | Descending motor pathway(Domi, deVeber et al. 2020) |
| Posterior limb of the internal capsule (PLIC) L/R | Motor-control white matter connectivity(Ghazi Sherbaf, Mojtahed Zadeh et al. 2019) |
| Anterior limb of the internal capsule (ALIC) L/R | Motor-control white matter connectivity(Yu, Chen et al. 2022) |
| Anterior, superior, and posterior corona radiata (ACR/SCR/PCR) L/R | Motor-control white matter connectivity(Li, Tan et al. 2026) |
| Medial lemniscus L/R | Somatosensory circuit(Jang, Kwon et al. 2012) |
| Posterior thalamic radiation (PTR) L/R | Thalamo-cortical sensorimotor circuit(Linortner, McDaniel et al. 2020) |
| Superior longitudinal fasciculus (SLF) L/R | Motor planning(Janelle, Iorio-Morin et al. 2022) |
| External capsule L/R | Motor-control white matter connectivity(Yu, Chen et al. 2022) |
| Sagittal stratum L/R | White matter connectivity(Haghshomar, Dolatshahi et al. 2018) |
| Genu, body, splenium, and tapetum of the corpus callosum (GCC/BCC/SCC/Tapetum) L/R | Interhemispheric connectivity(Amandola, Sinha et al. 2022) |
| Cingulum precentral L/R | White matter connectivity(Kollenburg, Arnts et al. 2025) |
| Cingulum postcentral L/R | White matter connectivity(Kollenburg, Arnts et al. 2025) |

**Table S4. Condensed summary of the 47 functional magnetic resonance imaging regions of interest selected for relevance to Parkinson's disease motor scores.**

A total of 47 Parkinson's disease- and motor-relevant functional MRI regions of interest were selected from the Schaefer et al. 2018 atlas(Schaefer, Kong et al. 2018) with subcortical and cerebellar labels(Nguyen, Raval et al. 2021). Selected ROIs included striatal, thalamic, precentral, postcentral, cerebellar, insular, and precuneus parcels with relevance to basal ganglia, thalamo-cortical, sensorimotor, cerebellar, and distributed cortical networks that may contribute to motor dysfunction in Parkinson's disease. For simplicity, bilateral, network-specific, and anatomically related ROIs are presented as grouped entries, while the selected feature set comprises 47 individual Schaefer et al 2018 atlas with subcortical structures derived features. Left and right labels indicate hemisphere.

| ROI(s) | Impact on the motor circuit |
|---|---|
| Visual Network Striatum L | Basal ganglia motor circuit(Zhai, Tanimura et al. 2018) |
| Somatomotor Network Striatum L/R | Basal ganglia motor circuit(Zhai, Tanimura et al. 2018) |
| Dorsal Attention Network Striatum L/R | Basal ganglia motor circuit(Zhai, Tanimura et al. 2018) |
| Salience Network Striatum L/R | Basal ganglia motor circuit(Zhai, Tanimura et al. 2018) |
| Limbic Network Striatum L/R | Basal ganglia motor circuit(Zhai, Tanimura et al. 2018) |
| Executive Control Network Striatum L/R | Basal ganglia motor circuit(Zhai, Tanimura et al. 2018) |
| Default Mode Network Striatum L/R | Basal ganglia motor circuit(Zhai, Tanimura et al. 2018) |
| Thalamus L/R | Thalamo-cortical / basal ganglia–cerebellar integration(Bosch-Bouju, Hyland et al. 2013) |
| Somatomotor Network Precentral 1–3 L/R | Primary motor cortex(Delgado-Alvarado, Ferrer-Gallardo et al. 2023) |
| Dorsal Attention Network Precentral L | Primary motor cortex(Delgado-Alvarado, Ferrer-Gallardo et al. 2023) |
| Somatomotor Network Postcentral L/R | somatosensory cortex(Delgado-Alvarado, Ferrer-Gallardo et al. 2023) |
| Dorsal Attention Network Postcentral L/R | somatosensory cortex(Delgado-Alvarado, Ferrer-Gallardo et al. 2023) |
| Cerebellum 1–7 L/R | Cerebellum(Wu and Hallett 2013, Guell, Schmahmann et al. 2018) |
| Salience Network Insular L/R | Sensorimotor integration(Criaud, Christopher et al. 2016, Tinaz, Para et al. 2018) |
| Somatomotor Network Insula R | Sensorimotor integration(Tinaz, Para et al. 2018) |
| Default Mode Network Precuneus 1–2 R | Default network(Criaud, Christopher et al. 2016, Ruppert, Greuel et al. 2021) |
| Dorsal Attention Network Precuneus R | Attention network(Boord, Madhyastha et al. 2017) |

## 2.5. Supplementary Figures

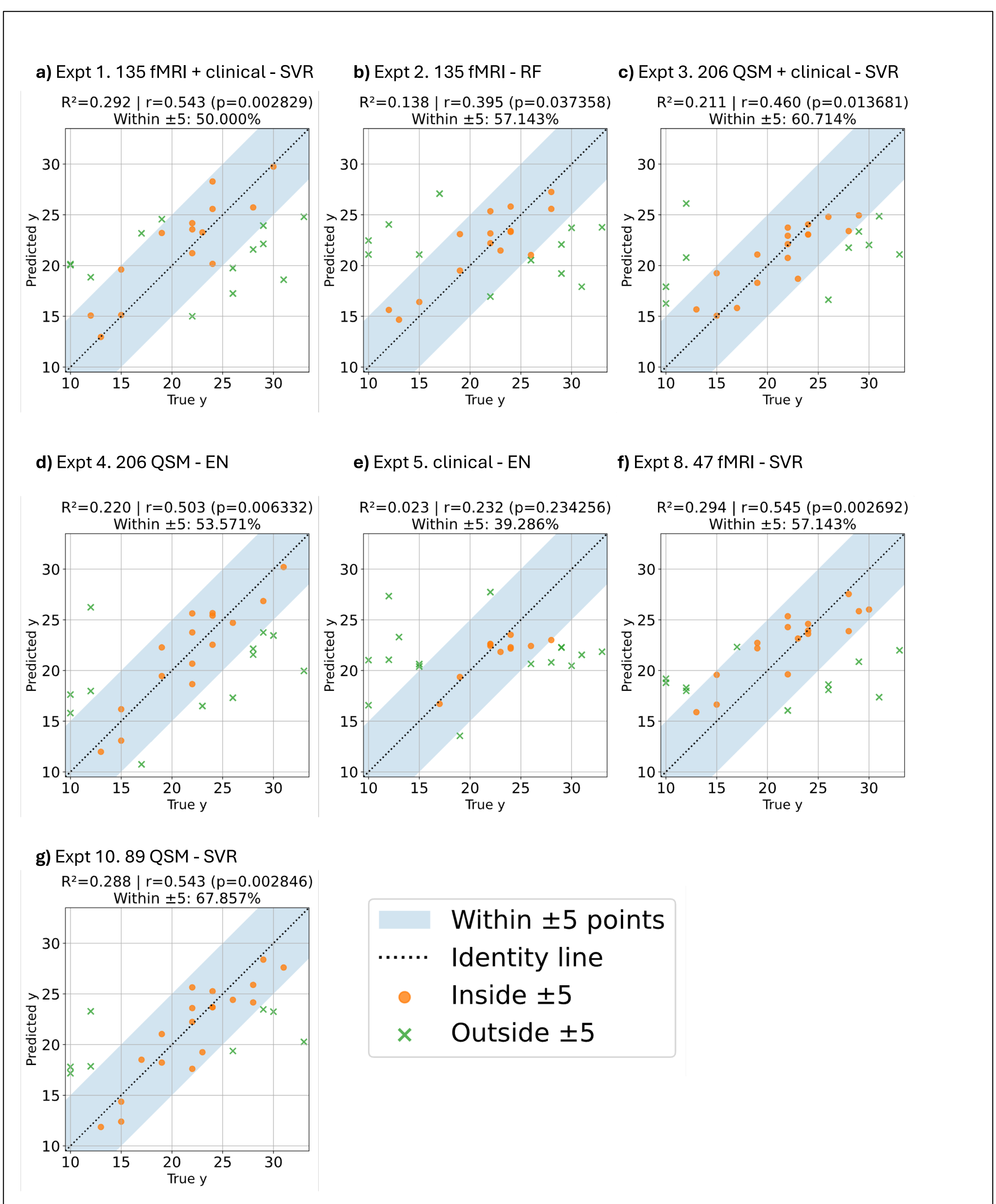

**Fig. S1. Nested cross-validated $R^2$ on held-out data (5 outer folds and 4 inner folds for hyperparameter optimization).**

Top-performing models among the four regression models evaluated. The target was MDS-UPDRS Part III, predicted from different combinations of clinical, fMRI, and QSM features. a) Expt 1. 135 fMRI+clinical. b) Expt 2. 135 fMRI c) Expt 3. 206 QSM+clinical d) Expt 4. 206 QSM e) Expt 5. clinical. f) Expt 8. 47 fMRI. g) Expt 10. 89 QSM.

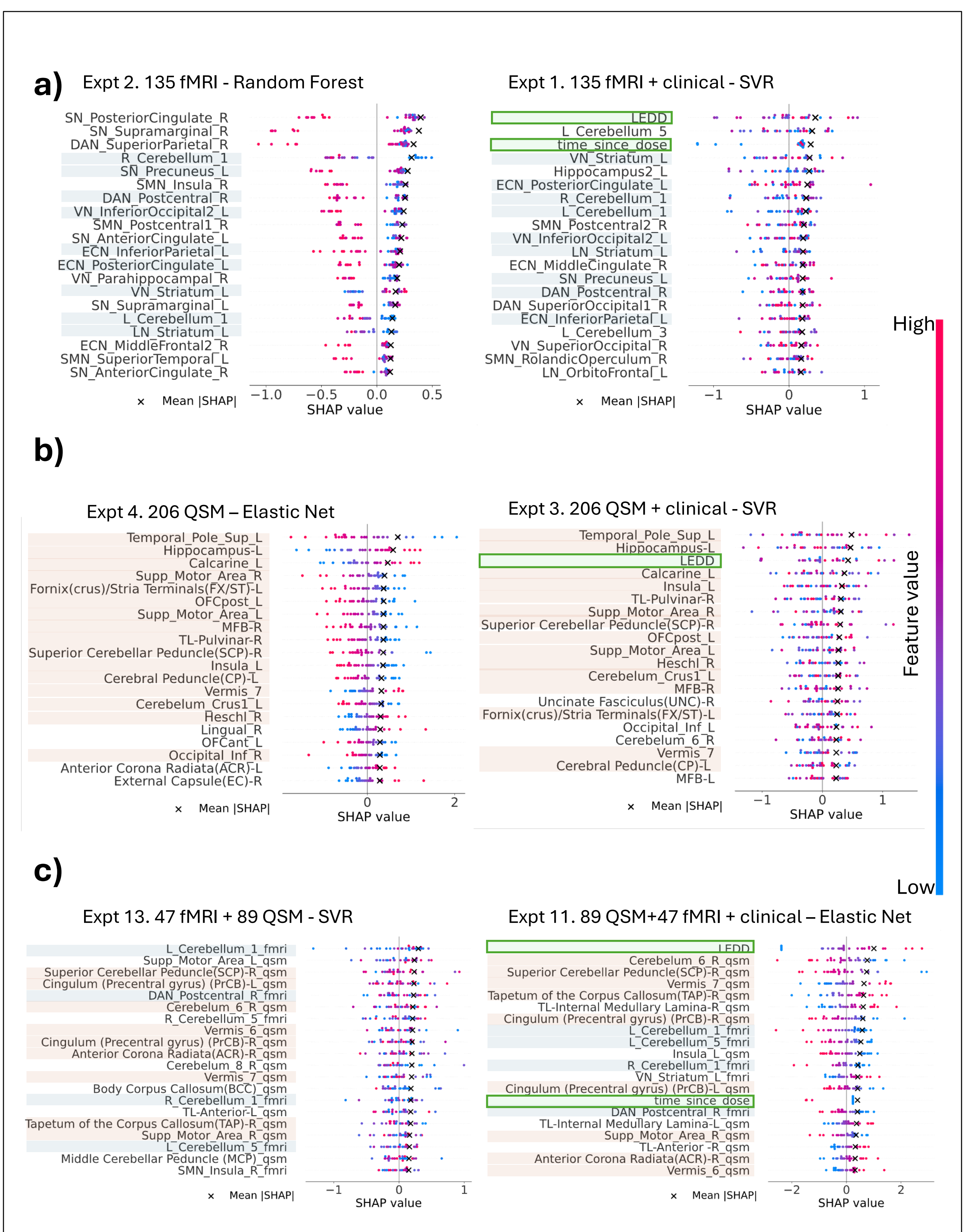

**Fig. S2-part1. Top 20 Features that most strongly influence MDS-UPDRS part 3 prediction from different combinations of QSM, fMRI and clinical data.**

Per-region SHAP distributions across subjects, with red/blue color encoding high/low QSM values and X indicating mean absolute SHAP. Each point is one subject's SHAP value for that region (positive = increases MDS-UPDRS, negative = lowers it). **a)** Expt 1. fMRI+clinical. and Expt 2. fMRI **b)** Expt 3. QSM+clinical and Expt 4. QSM **c)** Expt 11. 89 QSM+47 fMRI+clinical and Expt 13. 47 fMRI + 89 QSM. For a and b, blue shade corresponds to matching fMRI features and orange shade corresponds to matching QSM features within each panel. Green box highlights clinical features.

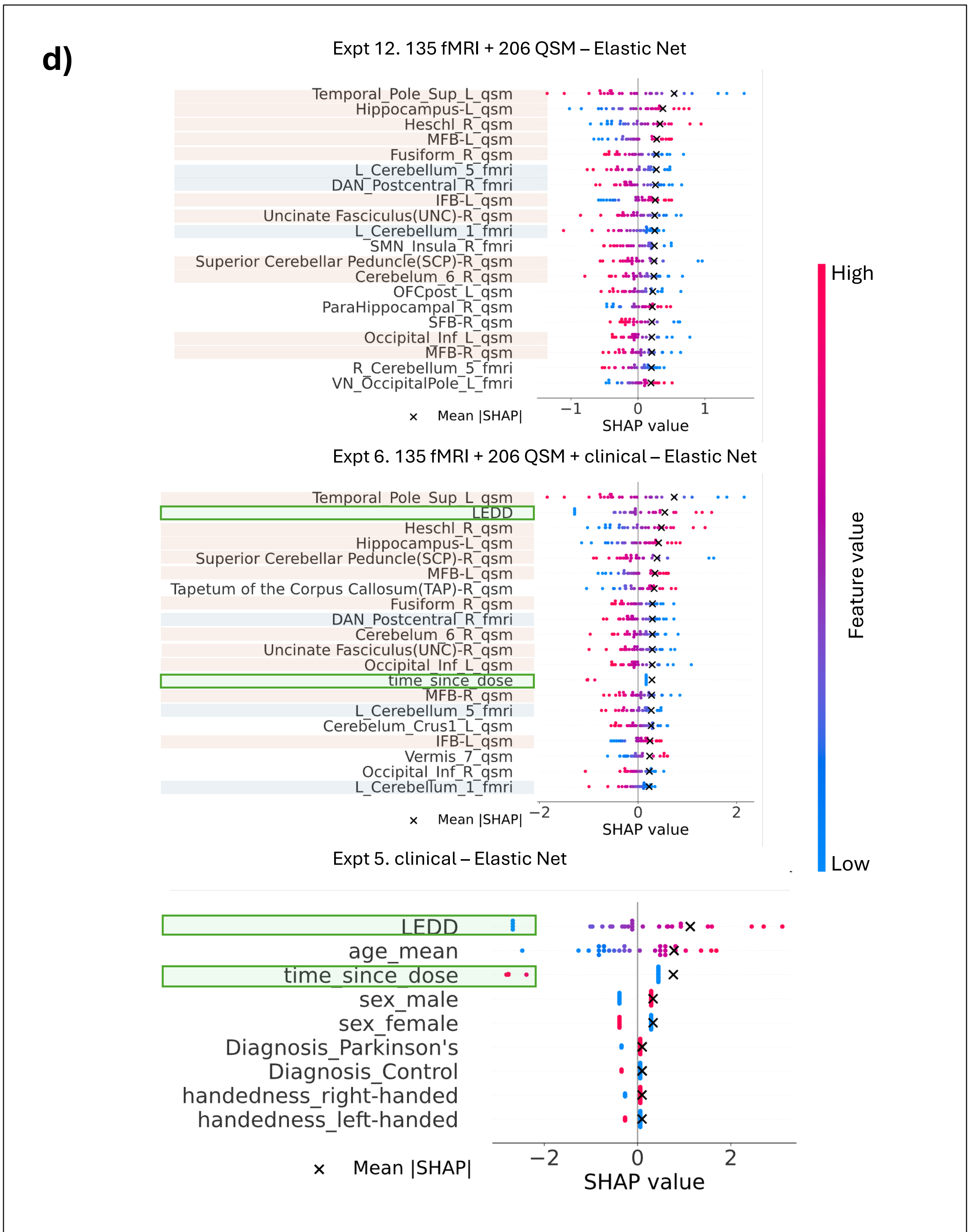

**Fig. S2-part2. Top 20 Features that most strongly influence MDS-UPDRS part 3 prediction from different combinations of QSM, fMRI and clinical data.**

Per-region SHAP distributions across subjects, with red/blue color encoding high/low QSM values and X indicating mean absolute SHAP. Each point is one subject's SHAP value for that region (positive = increases MDS-UPDRS, negative = lowers it). **d)** Expt 12. 135 fMRI + 206 QSM, Expt 6. fMRI+ QSM+clinical and Expt 5. clinical. Blue shade corresponds to matching fMRI features and orange shade corresponds to matching QSM features, green box highlights matching clinical features.

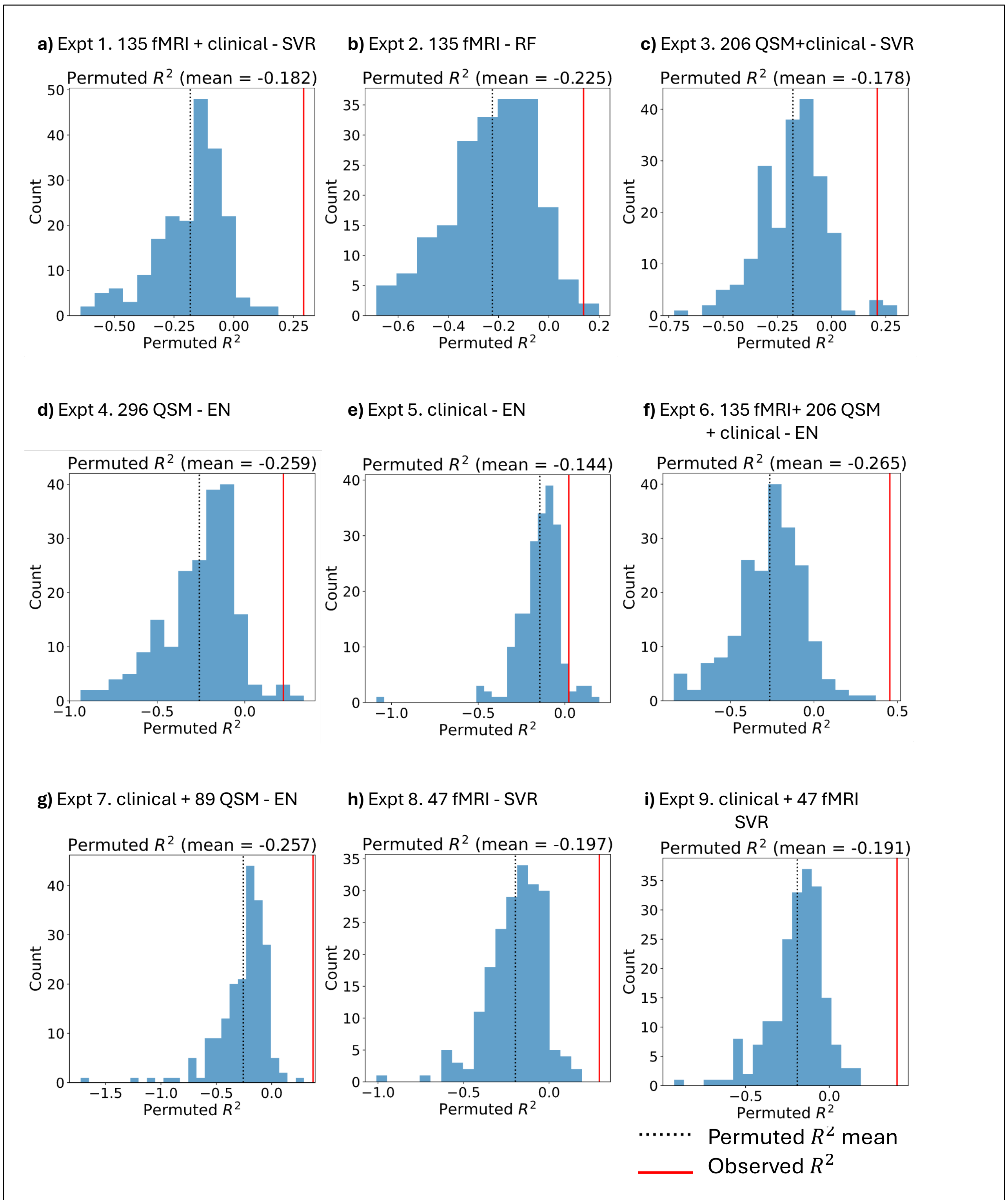

**Fig. S3-part1. Full nested cross-validation permutation test distributions for the selected winning models across experiments.**

For each model, the entire nested cross-validation procedure, including inner-fold hyperparameter optimization, was rerun under 200 random permutations of the outcome labels. Histograms show the null distribution of pooled held-out $R^2$values, the red vertical line marks the observed pooled held-out $R^2$, and the dotted black line marks the permutation mean. **a)** Expt 1. fMRI+clinical **b)** Expt 2. fMRI **c)** Expt 3. QSM+clinical **d)** Expt 4. QSM **e)** Expt 5. clinical **f)** Expt 6. fMRI+ QSM+clinical **g)** Expt 7. clinical + 89 QSM **h)** Expt 8. 47 fMRI **i)** Expt 9. clinical + 47 fMRI.

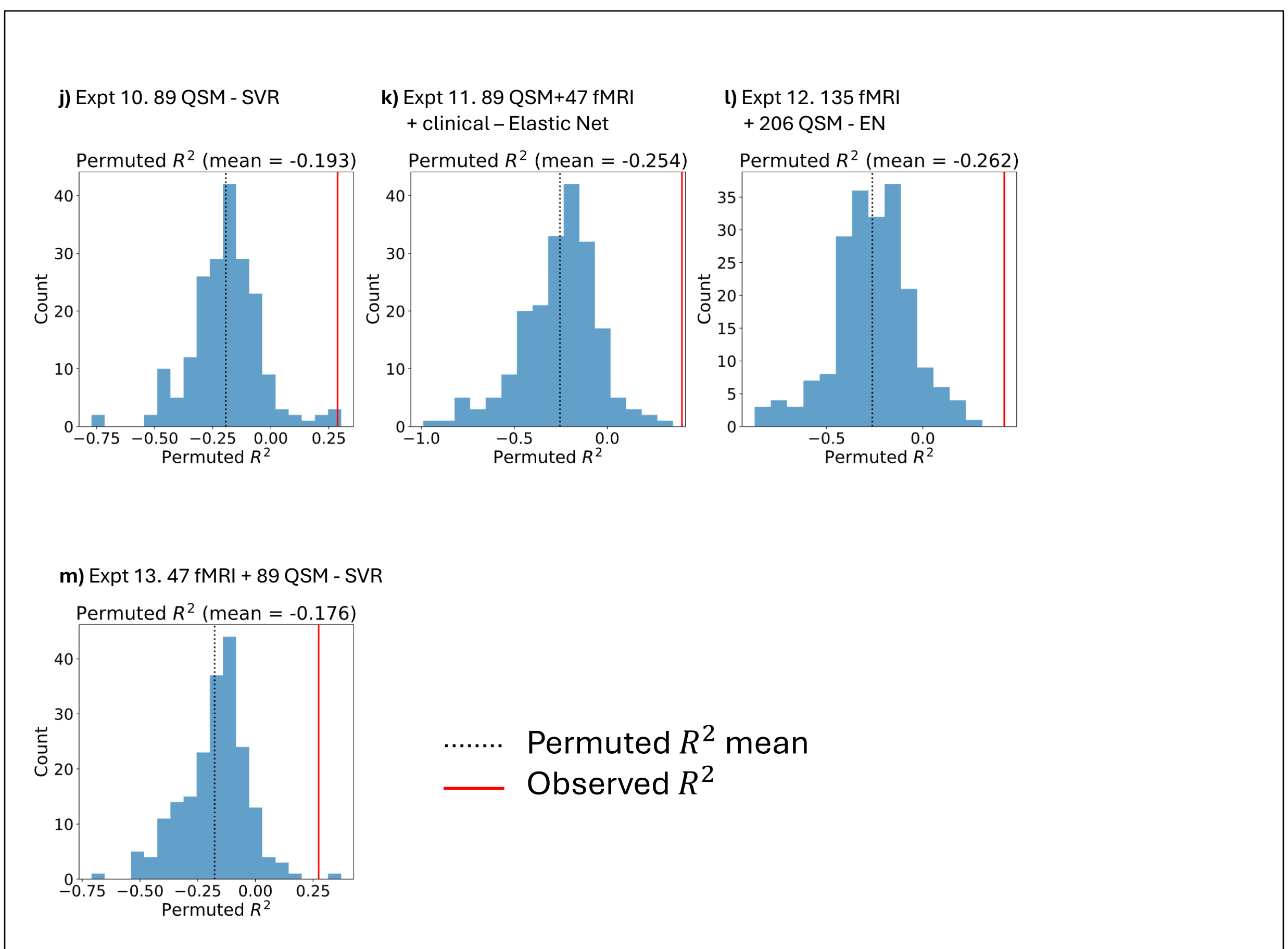

**Fig. S3-part2. Full nested cross-validation permutation test distributions for the selected winning models across experiments.**

For each model, the entire nested cross-validation procedure, including inner-fold hyperparameter optimization, was rerun under 200 random permutations of the outcome labels. Histograms show the null distribution of pooled held-out $R^2$values, the red vertical line marks the observed pooled held-out $R^2$ ,and the dotted black line marks the permutation mean. **j)** Expt 10. 89 QSM **k)** Expt 11. 89 QSM+47 fMRI+clinical **l)** Expt 12. 135 fMRI + 206 QSM and **m)** Expt 13. 47 fMRI + 89 QSM.

# References

Abraham, A., F. Pedregosa, M. Eickenberg, P. Gervais, A. Mueller, J. Kossaifi, A. Gramfort, B. Thirion and G. Varoquaux (2014). "Machine learning for neuroimaging with scikit-learn." Front Neuroinform **8**: 14.
Amandola, M., A. Sinha, M. J. Amandola and H. C. Leung (2022). "Longitudinal corpus callosum microstructural decline in early-stage Parkinson's disease in association with akinetic-rigid symptom severity." NPJ Parkinsons Dis **8**(1): 108.
Avants, B. B., N. J. Tustison, G. Song, P. A. Cook, A. Klein and J. C. Gee (2011). "A reproducible evaluation of ANTs similarity metric performance in brain image registration." Neuroimage **54**(3): 2033-2044.
Basile, G. A., M. Quartu, S. Bertino, M. P. Serra, M. Boi, A. Bramanti, G. P. Anastasi, D. Milardi and A. Cacciola (2021). "Red nucleus structure and function: from anatomy to clinical neurosciences." Brain Struct Funct **226**(1): 69-91.
Boord, P., T. M. Madhyastha, M. K. Askren and T. J. Grabowski (2017). "Executive attention networks show altered relationship with default mode network in PD." Neuroimage Clin **13**: 1-8.
Borich, M. R., S. M. Brodie, W. A. Gray, S. Ionta and L. A. Boyd (2015). "Understanding the role of the primary somatosensory cortex: Opportunities for rehabilitation." Neuropsychologia **79**(Pt B): 246-255.
Bosch-Bouju, C., B. I. Hyland and L. C. Parr-Brownlie (2013). "Motor thalamus integration of cortical, cerebellar and basal ganglia information: implications for normal and parkinsonian conditions." Front Comput Neurosci **7**: 163.
Chu, H. Y., Y. Smith, W. W. Lytton, S. Grafton, R. Villalba, G. Masilamoni and T. Wichmann (2024). "Dysfunction of motor cortices in Parkinson's disease." Cereb Cortex **34**(7).
Cox, R. W. (1996). "AFNI: software for analysis and visualization of functional magnetic resonance neuroimages." Comput Biomed Res **29**(3): 162-173.
Craddock, C., S. Sikka, B. Cheung, R. Khanuja, S. Ghosh, C. Yan, Q. Li, D. Lurie, J. Vogelstein, R. Burns, S. Colcombe, M. Mennes, C. Kelly, A. Di Martino, F. Castellanos and M. Milham (2013). "Towards Automated Analysis of Connectomes: The Configurable Pipeline for the Analysis of Connectomes (C-PAC)." Frontiers in Neuroinformatics **7**.
Criaud, M., L. Christopher, P. Boulinguez, B. Ballanger, A. E. Lang, S. S. Cho, S. Houle and A. P. Strafella (2016). "Contribution of insula in Parkinson's disease: A quantitative meta-analysis study." Hum Brain Mapp **37**(4): 1375-1392.
Delgado-Alvarado, M., V. J. Ferrer-Gallardo, P. M. Paz-Alonso, C. Caballero-Gaudes and M. C. Rodriguez-Oroz (2023). "Interactions between functional networks in Parkinson's disease mild cognitive impairment." Sci Rep **13**(1): 20162.
Domi, T., G. deVeber, D. Mikulis and A. Kassner (2020). "Wallerian Degeneration of the Cerebral Peduncle and Association with Motor Outcome in Childhood Stroke." Pediatr Neurol **102**: 67-73.
DuPre, E., T. Salo, Z. Ahmed, P. Bandettini, K. Bottenhorn, C. Caballero-Gaudes, L. Dowdle, J. Gonzalez-Castillo, S. Heunis, P. Kundu, A. Laird, R. Markello, C. Markiewicz, S. Moia, I. Staden, J. Teves, E. Uruñuela, M. Vaziri-Pashkam, K. Whitaker and D. Handwerker (2021). "TE-dependent analysis of multi-echo fMRI with tedana." Journal of Open Source Software **6**(66).
Fonov, V. S., A. C. Evans, R. C. McKinstry, C. R. Almli and D. L. Collins (2009). "Unbiased nonlinear average age-appropriate brain templates from birth to adulthood." NeuroImage **47**.

Galvan, A., A. Devergnas and T. Wichmann (2015). "Alterations in neuronal activity in basal ganglia-thalamocortical circuits in the parkinsonian state." Front Neuroanat **9**: 5.
Ghazi Sherbaf, F., M. Mojtahed Zadeh, M. Haghshomar and M. H. Aarabi (2019). "Posterior limb of the internal capsule predicts poor quality of life in patients with Parkinson's disease: connectometry approach." Acta Neurol Belg **119**(1): 95-100.
Gorgolewski, K., C. D. Burns, C. Madison, D. Clark, Y. O. Halchenko, M. L. Waskom and S. S. Ghosh (2011). "Nipype: a flexible, lightweight and extensible neuroimaging data processing framework in python." Front Neuroinform **5**: 13.
Grahn, J. A., J. A. Parkinson and A. M. Owen (2008). "The cognitive functions of the caudate nucleus." Prog Neurobiol **86**(3): 141-155.
Guell, X., J. D. Schmahmann, J. Gabrieli and S. S. Ghosh (2018). "Functional gradients of the cerebellum." Elife **7**.
Guo, W., D. Zhang, J. Sun, L. Chen, T. Wu and E. Xu (2024). "Quantitative susceptibility mapping of subcortical iron deposition in Parkinson disease and multiple system atrophy: clinical correlations and diagnostic implications." Quant Imaging Med Surg **14**(7): 4464-4474.
Haghshomar, M., M. Dolatshahi, F. Ghazi Sherbaf, H. Sanjari Moghaddam, M. Shirin Shandiz and M. H. Aarabi (2018). "Disruption of Inferior Longitudinal Fasciculus Microstructure in Parkinson's Disease: A Systematic Review of Diffusion Tensor Imaging Studies." Front Neurol **9**: 598.
He, N., P. Huang, H. Ling, J. Langley, C. Liu, B. Ding, J. Huang, H. Xu, Y. Zhang, Z. Zhang, X. Hu, S. Chen and F. Yan (2017). "Dentate nucleus iron deposition is a potential biomarker for tremor-dominant Parkinson's disease." NMR Biomed **30**(4).
Hoffstaedter, F., C. Grefkes, S. Caspers, C. Roski, N. Palomero-Gallagher, A. R. Laird, P. T. Fox and S. B. Eickhoff (2014). "The role of anterior midcingulate cortex in cognitive motor control: evidence from functional connectivity analyses." Hum Brain Mapp **35**(6): 2741-2753.
Huang, Y., L. Chen, X. Li and J. Liu (2024). "Improved test-retest reliability of R2* and susceptibility quantification using multishot multi-echo 3D EPI." Magn Reson Med **91**(6): 2310-2319.
Janelle, F., C. Iorio-Morin, S. D'Amour and D. Fortin (2022). "Superior Longitudinal Fasciculus: A Review of the Anatomical Descriptions With Functional Correlates." Front Neurol **13**: 794618.
Jang, S. H., Y. H. Kwon, M. Y. Lee, D. Y. Lee and J. H. Hong (2012). "Termination differences in the primary sensorimotor cortex between the medial lemniscus and spinothalamic pathways in the human brain." Neurosci Lett **516**(1): 50-53.
Jang, S. H. and H. D. Lee (2020). "Relationship between ataxia and inferior cerebellar peduncle injury in patients with cerebral infarct." Medicine (Baltimore) **99**(9): e19344.
Jenkinson, M., P. Bannister, M. Brady and S. Smith (2002). "Improved Optimization for the Robust and Accurate Linear Registration and Motion Correction of Brain Images." NeuroImage **17**(2): 825-841.
Kimura, Y., S. Yamada, K. Komatsu, R. Enatsu, R. Sato, C. Kamada, A. Sasagawa, T. Hirano, M. Arihara and N. Mikuni (2024). "An anatomo-functional study of the interactivity between the paracentral lobule and the primary motor cortex." J Neurosurg **141**(4): 1096-1104.
Kollenburg, L., H. Arnts, A. Green, I. Strauss, S. Vinke and E. Kurt (2025). "The cingulum: anatomy, connectivity and what goes beyond." Brain Commun **7**(1): fcaf048.
Li, X., C. Tan, S. Cai, Q. Liu, M. Wang, Y. Liu, J. Yuan, Y. Tang, Y. Niu, C. Huang, W. He, Q. Shen and H. Liao (2026). "White matter alterations in right-onset versus left-onset Parkinson's disease in an early stage." Behav Brain Res **505**: 116116.

Linortner, P., C. McDaniel, M. Shahid, T. F. Levine, L. Tian, B. Cholerton and K. L. Poston (2020). "White Matter Hyperintensities Related to Parkinson's Disease Executive Function." Mov Disord Clin Pract **7**(6): 629-638.
Maliia, M. D., C. Donos, A. Barborica, I. Popa, J. Ciurea, S. Cinatti and I. Mindruta (2018). "Functional mapping and effective connectivity of the human operculum." Cortex **109**: 303-321.
McGregor, M. M. and A. B. Nelson (2019). "Circuit Mechanisms of Parkinson's Disease." Neuron **101**(6): 1042-1056.
Nguyen, K. P., V. Raval, A. Treacher, C. Mellema, F. F. Yu, M. C. Pinho, R. M. Subramaniam, R. B. Dewey, Jr. and A. A. Montillo (2021). "Predicting Parkinson's disease trajectory using clinical and neuroimaging baseline measures." Parkinsonism Relat Disord **85**: 44-51.
Park, I. S., N. J. Lee and I. J. Rhyu (2018). "Roles of the Declive, Folium, and Tuber Cerebellar Vermian Lobules in Sportspeople." J Clin Neurol **14**(1): 1-7.
Pietracupa, S., A. Ojha, D. Belvisi, C. Piervincenzi, S. Tommasin, N. Petsas, M. I. De Bartolo, M. Costanzo, A. Fabbrini, A. Conte, A. Berardelli and P. Pantano (2024). "Understanding the role of cerebellum in early Parkinson's disease: a structural and functional MRI study." NPJ Parkinsons Dis **10**(1): 119.
Rahimpour, S., S. Rajkumar and M. Hallett (2022). "The Supplementary Motor Complex in Parkinson's Disease." J Mov Disord **15**(1): 21-32.
Riederer, P., T. Nagatsu, M. B. H. Youdim, M. Wulf, J. M. Dijkstra and J. Sian-Huelsmann (2023). "Lewy bodies, iron, inflammation and neuromelanin: pathological aspects underlying Parkinson's disease." J Neural Transm (Vienna) **130**(5): 627-646.
Ruppert, M. C., A. Greuel, J. Freigang, M. Tahmasian, F. Maier, J. Hammes, T. van Eimeren, L. Timmermann, M. Tittgemeyer, A. Drzezga and C. Eggers (2021). "The default mode network and cognition in Parkinson's disease: A multimodal resting-state network approach." Hum Brain Mapp **42**(8): 2623-2641.
Schaefer, A., R. Kong, E. M. Gordon, T. O. Laumann, X. N. Zuo, A. J. Holmes, S. B. Eickhoff and B. T. T. Yeo (2018). "Local-Global Parcellation of the Human Cerebral Cortex from Intrinsic Functional Connectivity MRI." Cereb Cortex **28**(9): 3095-3114.
Seo, J. P. and S. H. Jang (2013). "Characteristics of corticospinal tract area according to pontine level." Yonsei Med J **54**(3): 785-787.
Smith, S. M. (2002). "Fast robust automated brain extraction." Hum Brain Mapp **17**(3): 143-155.
Tamanini, J. V. G., G. A. S. Ribeiro, A. T. Kimura, L. F. Borella, T. A. Freddi and F. Reis (2024). "Hyperintense lesions of the middle cerebellar peduncle and beyond: a pictorial essay." Radiol Bras **57**: e20240001.
Tinaz, S., K. Para, A. Vives-Rodriguez, V. Martinez-Kaigi, K. Nalamada, M. Sezgin, D. Scheinost, M. Hampson, E. D. Louis and R. T. Constable (2018). "Insula as the Interface Between Body Awareness and Movement: A Neurofeedback-Guided Kinesthetic Motor Imagery Study in Parkinson's Disease." Front Hum Neurosci **12**: 496.
van Gelderen, P., X. Li, J. A. de Zwart, E. S. Beck, S. V. Okar, Y. Huang, K. Lai, J. Sulam, P. C. M. van Zijl, D. S. Reich, J. H. Duyn and J. Liu (2023). "Effect of motion, cortical orientation and spatial resolution on quantitative imaging of cortical R(2)* and magnetic susceptibility at 0.3 mm in-plane resolution at 7 T." Neuroimage **270**: 119992.
Wu, T. and M. Hallett (2013). "The cerebellum in Parkinson's disease." Brain **136**(Pt 3): 696-709.

Xu, D., Q. Ding and H. Wang (2020). "Corticospinal Tract Impairment of Patients With Parkinson's Disease: Triple Stimulation Technique Findings." Front Aging Neurosci **12**: 588085.
Yu, J., L. Chen, G. Cai, Y. Wang, X. Chen, W. Hong and Q. Ye (2022). "Evaluating white matter alterations in Parkinson's disease-related parkin S/N167 mutation carriers using tract-based spatial statistics." Quant Imaging Med Surg **12**(8): 4272-4285.
Zang, Y., T. Jiang, Y. Lu, Y. He and L. Tian (2004). "Regional homogeneity approach to fMRI data analysis." Neuroimage **22**(1): 394-400.
Zhai, S., A. Tanimura, S. M. Graves, W. Shen and D. J. Surmeier (2018). "Striatal synapses, circuits, and Parkinson's disease." Curr Opin Neurobiol **48**: 9-16.
Zhang, X., R. Li, Y. Xia, H. Zhao, L. Cai, J. Sha, Q. Xiao, J. Xiang, C. Zhang and K. Xu (2022). "Topological patterns of motor networks in Parkinson's disease with different sides of onset: A resting-state-informed structural connectome study." Front Aging Neurosci **14**: 1041744.
Zhang, Y., H. Wei, M. J. Cronin, N. He, F. Yan and C. Liu (2018). "Longitudinal atlas for normative human brain development and aging over the lifespan using quantitative susceptibility mapping." Neuroimage **171**: 176-189.